\documentclass{article}

\usepackage{amsmath,graphicx,algorithm,algorithmic,xspace,amsfonts,amsthm}
\usepackage{amsthm}
\usepackage{subfigure}
\usepackage{multicol}
\usepackage{tikz,spconf}

\title{Combined Modeling of Sparse and Dense Noise for Improvement of Relevance Vector Machine}
\name{Martin~Sundin, Saikat~Chatterjee and Magnus~Jansson\thanks{This work was partially supported by the Swedish Research Council under contract 621-2011-5847.}}
\address{ACCESS Linnaeus Center, School of Electrical Engineering\\
KTH Royal Institute of Technology, Stockholm, Sweden\\
{\small \tt masundi@kth.se, sach@kth.se, janssonm@kth.se}}

\begin{document}


\maketitle

\begin{abstract}
Using a Bayesian approach, we consider the problem of recovering sparse signals under additive sparse and dense noise. Typically, sparse noise models outliers, impulse bursts or data loss. To handle sparse noise, existing methods simultaneously estimate the sparse signal of interest and the sparse noise of no interest. For estimating the sparse signal, without the need of estimating the sparse noise, we construct a robust Relevance Vector Machine (RVM). In the RVM, sparse noise and ever present dense noise are treated through a combined noise model. The precision of combined noise is modeled by a diagonal matrix. We show that the new RVM update equations correspond to a non-symmetric sparsity inducing cost function. Further, the combined modeling is found to be computationally more efficient. We also extend the method to block-sparse signals and noise with known and unknown block structures. Through simulations, we show the performance and computation efficiency of the new RVM in several applications: recovery of sparse and block sparse signals, housing price prediction and image denoising.

\end{abstract}


%

\section{Introduction}
\label{sec:intro}

Noise modeling has an important role in the Bayesian inference setup to achieve 
better robustness and accuracy. 
Typically noise is considered to be additive and dense (possibly even white) in nature. In this paper we investigate the effect of sparse noise modeling 
in a standard Bayesian inference tool called the \emph{Relevance 
Vector Machine} (RVM) \cite{Tipping1}. 

The RVM is a Bayesian sparse kernel technique for applications in regression and classification \cite{Tipping1}.
Interest in the RVM can be attributed to the cause that it shares many characteristics
of the popular support vector machine whilst providing Bayesian advantages \cite{Bishop,Wipf,Wipf2}, mainly 
providing posteriors for the object of interest. Generally the RVM is a fully Bayesian technique that 
aims to learn all the relevant system parameters iteratively to 
infer the object of interest. In a linear model setup used for regression, 
RVM introduces sparsity through a weight vector where the weights are essential to form 
linear combinations of relevant kernels to predict the object of interest; the weight vector
is a set of system parameters and its sparsity leads to reduction of model complexity
for regression. 
Naturally, the RVM has been further used
for sparse representation techniques as well as developing Bayesian
compressive sensing methods \cite{BayesianCS}. 

For a Bayesian linear model, the standard RVM uses a multivariate isotropic Gaussian prior 
to model the additive dense noise. Here isotropic means that the associated covariance matrix
is proportional to the identity matrix. Such a dense noise model has inherent limitations to accommodate
instances of outliers \cite{Mitra2,Laska,Jin,Vehkapera,Cherian}, impulse bursts \cite{Giri,Giacobello} or missing (lost) data \cite{Kekatos,Carrillo}. We hypothesize that a sparse and dense noise
model can accommodate for the statistics of a variety of noise types,
without causing degradation in performance for any noise type compared to the standard case
of using only a dense noise model. In this paper, we develop RVM for such a combined (joint)
sparse and dense noise scenario.

%
%
%

\subsection{System model}

We consider the following linear system model
\begin{align}
\label{eq:sparsenoise}
\mathbf{y} = \mathbf{A}\mathbf{x} + \mathbf{e} + \mathbf{n} , 
\end{align}
where $\mathbf{y}\in\mathbb{R}^m$ is the measurements, $\mathbf{x}\in\mathbb{R}^n$ is a 
sparse vector (for example weights in regression or sparse signal to estimate in compressed sensing),
$\mathbf{A} \in \mathbb{R}^{m \times n}$ is a known system matrix (for 
example, regressors or sampling system). Further, $\mathbf{e}\in\mathbb{R}^m$ is 
sparse noise and $\mathbf{n}\in\mathbb{R}^m$ is dense noise. Using $\ell_0$-norm notation
to represent the number of non-zeros in a vector, we assume that $||\mathbf{x}||_0 \ll n$ and 
$||\mathbf{e}||_0 \ll m$ are small and unknown. 
The random vectors $\mathbf{x}$, $\mathbf{e}$ and $\mathbf{n}$ are independent.
The model \eqref{eq:sparsenoise} is used in face recognition 
\cite{Wright}, image denoising \cite{Mitra2} and compressed sensing \cite{BayesianCS}.

\subsection{Our contribution}

We develop a RVM for the model \eqref{eq:sparsenoise}, by treating $\mathbf{e} + \mathbf{n}$
as a combined noise. By learning parameters of $\mathbf{x}$ and $\mathbf{e} + \mathbf{n}$,
we estimate $\mathbf{x}$ without the need of estimating $\mathbf{e}$. 
We also consider the scenario where the signal $\mathbf{x}$ and noise $\mathbf{e}$ are block sparse. By using techniques similar to the ones in \cite{zhang} we generalize the methods to signals with unknown block structure. 
The main technical contribution is to derive update equations 
that are used iteratively for estimation of parameters in the new RVM. We refer to the new RVM as the
RVM for combined sparse and dense noise (SD-RVM). By an approximate analysis, the SD-RVM algorithm 
is shown to be equivalent to the minimization of a non-symmetric sparsity inducing cost function. 
Finally, the performance of SD-RVM is 
evaluated numerically using examples from compressed sensing, block sparse signal recovery, house price prediction and image denoising.
Throughout the paper, we take an approach of comparing SD-RVM vis-a-vis the existing Robust Bayesian RVM (RB-RVM) \cite{Mitra2} (described in the next section).

\subsection{Prior work}

To establish relevance of our work we briefly describe prior work in this section. 
Almost all prior works \cite{Mitra2,Laska,Jin,Vehkapera} translate the linear setup \eqref{eq:sparsenoise} to the equivalent setup
\begin{align}
\label{eq:equivalent_sparsenoise}
\mathbf{y} = 
 \left[ \begin{array}{cc}
        \mathbf{A} & \mathbf{I}_{m}
       \end{array}
 \right] 
 \left[ \begin{array}{c}
        \mathbf{x} \\ \mathbf{e}
       \end{array}
 \right] 
 + \mathbf{n},
\end{align}
where $\mathbf{I}_m$ is the $m \times m$ identity matrix, $\left[ \begin{array}{cc} \mathbf{A} & \mathbf{I}_{m} \end{array}  \right] $ acts as the effective
system matrix and $\left[ \mathbf{x}^{\top} \,\, \mathbf{e}^{\top} \right]^{\top}$ acts as the parameter vector to be estimated.
The RB-RVM of \cite{Mitra2} uses the standard RVM approach 
for \eqref{eq:equivalent_sparsenoise}
directly. Hence RB-RVM learns model parameters for all three signals $\mathbf{x}=[x_1, \, x_2, \, \ldots, x_n]^{\top}$, 
$\mathbf{e}=[e_1, \, e_2, \, \ldots, e_m]^{\top}$ and $\mathbf{n}$, and
thus estimates both $\mathbf{x}$ and $\mathbf{e}$ jointly. 
RB-RVM assumes Gaussian priors
\begin{align*}
\mathbf{x} \sim \prod_{i=1}^{n} \mathcal{N}( 0,\gamma_i^{-1}), \,\, 
\mathbf{e} \sim \prod_{i=1}^{m} \mathcal{N}( 0,\nu_i^{-1}), \,\, \mathbf{n} \sim \mathcal{N}(\mathbf{0},\beta^{-1} \mathbf{I}_m),
\end{align*}
where the precisions (inverse variances) $\alpha_i$, $\nu_i$ and $\beta$ are unknown. 
The precisions are given Gamma priors
\begin{align}
p(\gamma_i) &= \mathrm{Gamma}(\gamma_i| a+1,b), \label{alphaprior}\\ 
p(\nu_i) &= \mathrm{Gamma}(\nu_i| a+1,b), \nonumber\\
p(\beta) &= \mathrm{Gamma}(\beta | c+1,d) , \nonumber
\end{align}
where $\mathrm{Gamma}(\gamma_i| a+1, b ) \propto \gamma_i^{a} e^{-b \gamma_i}$ \cite{Tipping1}.
Typical practice is to maximize $p(\mathbf{y} |\boldsymbol{\gamma},\boldsymbol{\nu},\beta)$ to infer the precisions, where we used boldface symbols to denote the vectors
\begin{align*}
&\boldsymbol{\gamma} = [\gamma_1,\gamma_2,\dots , \gamma_n]^\top ,\\
&\boldsymbol{\nu} = [\nu_1,\nu_2, \dots , \nu_m]^\top .
\end{align*}
Instead we take the alternative (full Bayesian) approach of maximizing $p(\mathbf{y},\boldsymbol{\gamma},\boldsymbol{\nu},\beta)$ and
assume that precisions have non-informative prior by taking the limit $( a,b,c,d ) \to \mathbf{0}$. For the distributions considered here, maximizing the conditional distribution $p(\mathbf{y} |\boldsymbol{\gamma},\boldsymbol{\nu},\beta)$ becomes equivalent to maximizing the joint distribution $p(\mathbf{y},\boldsymbol{\gamma},\boldsymbol{\nu},\beta)$, in  the limit of non-informative priors. In calculations, however, the parameters $( a,b,c,d )$ are often given small values to avoid numerical instabilities.
To estimate $\left[ \mathbf{x}^{\top} \,\, \mathbf{e}^{\top} \right]^{\top}$, RB-RVM fixes the precisions and sets
\begin{align}
&\left[ \hat{\mathbf{x}}^{\top} \,\, \hat{\mathbf{e}}^{\top} \right]^{\top} = \beta \boldsymbol{\Sigma}_{RB} [\mathbf{A}\,\ \mathbf{I}_{m}]^\top \mathbf{y} , \label{xupdate}\\
&\boldsymbol{\Sigma}_{RB} = \left(
\left(\begin{array}{cc}
\boldsymbol{\Gamma} & \mathbf{0}\\
\mathbf{0} & \mathbf{N}
\end{array} \right)
 + \beta  [\mathbf{A}\,\ \mathbf{I}_{m}]^{\top} 
[\mathbf{A}\,\ \mathbf{I}_{m}] \right)^{-1} , \nonumber
\end{align}
where $\boldsymbol{\Gamma} = \mathrm{diag}(\gamma_1,\gamma_2,\dots , \gamma_n)$ and $\mathbf{N} = \mathrm{diag}(\nu_1,\nu_2,\dots,\nu_m)$. 
The RB-RVM iteratively updates the precisions by maximizing  
$p(\mathbf{y},\boldsymbol{\gamma},\boldsymbol{\nu},\beta)$, resulting in the update equations
\begin{align}
&\gamma_i^{new} = \frac{1 - \gamma_i [\boldsymbol{\Sigma}_{RB}]_{ii}}{\hat{x}_i^2}, \,\,\,
\nu_i^{new} = \frac{1 - \nu_i [\boldsymbol{\Sigma}_{RB}]_{n+i,n+i}}{\hat{e}_i^2}, \nonumber\\
&\beta^{new} = \frac{\sum_{i=1}^n \gamma_i [\boldsymbol{\Sigma}_{RB}]_{ii} + \sum_{j=1}^m \nu_j [\boldsymbol{\Sigma}_{RB}]_{n+i,n+i} }{||\mathbf{y} - \mathbf{A}\mathbf{\hat{x}} - \mathbf{\hat{e}} ||_2^2} 
, \label{rvmbetaupdate}
\end{align}
where $[\boldsymbol{\Sigma}_{RB}]_{ii}$ denotes the $(i,i)$ component of the matrix $\boldsymbol{\Sigma}_{RB}$. 

The update equations \eqref{xupdate} and \eqref{rvmbetaupdate} are found by applying the standard RVM to \eqref{eq:equivalent_sparsenoise}. Derivations can be found in e.g. \cite{Tipping1,Bishop}. Iterating 
until convergence gives the final estimates $\hat{\mathbf{x}}$ and $\hat{\mathbf{e}}$.
In the iterations, some precisions become large, making their respective components 
in $\hat{\mathbf{x}}$ and $\hat{\mathbf{e}}$ close to zero. This makes the final estimate of $\hat{\mathbf{x}}$ and $\hat{\mathbf{e}}$
sparse.

RVM has high similarity with Sparse Bayesian Learning (SBL) \cite{zhang,zhang2,Wipf,Wipf2}. Sparse Bayesian learning has been used for structured sparse signals, for example block sparse signals \cite{zhang}, where the problem of unknown signal block structure was treated using overlapping blocks. The model extension of RB-RVM shown in \eqref{eq:equivalent_sparsenoise} for handling block sparse noise with unknown block structure is straight-forward to derive. However, in our formulation, as we are not estimating the noise explicitly, the use of block sparse noise with unknown block structure is non-trivial.

Further, non Bayesian (even not statistical) methods have been used for sparse estimation problems \cite{Laska,Chen2001,Lookahead,DIP}.
For example, the $\ell_1$-norm minimization method justice pursuit (JP) \cite{Laska} 
uses the optimization technique of the standard basis pursuit denoising method \cite{Chen2001}, as follows
\begin{align}
\mathbf{\hat{x},\hat{e}} = \arg \min_{\mathbf{x,e}} \, 
||\mathbf{x}||_1 + ||\mathbf{e}||_1  \label{jp}
\text{ s.t. } || \mathbf{y} - \mathbf{Ax} - \mathbf{e} ||_2 \leq \epsilon , 
\end{align}
where $\epsilon > 0$ is a model parameter set by the user. For unknown noise power,
it is impossible to know $\epsilon$ a-priori. We mention that a fully Bayesian setup
like the RVM does not require parameters set by a user.

\section{RVM for combined sparse and dense noise (SD-RVM)}

\subsection{SD-RVM Method}
\label{subsec:SD-RVM}
For \eqref{eq:sparsenoise}, we propose to use a combined model for the two additive noises, as follows 
\begin{align}
\label{robustbeta}
\mathbf{e + n} \sim \mathcal{N}(\mathbf{0},\mathbf{B}^{-1}) , 
\end{align}
where $\mathbf{B} = \mathrm{diag}(\beta_1, \beta_2 , \dots, \beta_m)$.
We also use $\boldsymbol{\beta}$ to denote the vector $\boldsymbol{\beta} = [\beta_1 , \beta_2, \dots , \beta_m]^\top$.
That means the two noises are treated as a single combined noise where each noise component has its own precision. 
The rationale is that we do not need to seperate the two noises. 
Although our model promotes sparsity in the noise we empirically find that it is able to model sparse and non-sparse noise.
Using the noise model \eqref{robustbeta} and that $\mathbf{x} \sim \prod_{i=1}^{n} \mathcal{N}( 0,\gamma_i^{-1})$, we find 
the maximum a posteriori (MAP) estimate
\begin{align*}
&\mathbf{\hat{x}} =  \boldsymbol{\Sigma} \mathbf{A}^\top \mathbf{B} \mathbf{y}, \nonumber\\
&\boldsymbol{\Sigma} = (\boldsymbol{\Gamma} + \mathbf{A}^\top \mathbf{B} \mathbf{A})^{-1}, \nonumber
\end{align*}
where as before $\boldsymbol{\Gamma} = \mathrm{diag}(\gamma_1, \gamma_2, \dots, \gamma_n)$. The precisions are updated as
\begin{align}
&\gamma_{i}^{new} = \frac{1 - \gamma_i \Sigma_{ii}}{\hat{x}_i^2}, \label{alphaupdate}\\
&\beta_j^{new} = \frac{1 - \beta_j [\mathbf{A}\boldsymbol{\Sigma} \mathbf{A}^\top]_{jj} }{[\mathbf{y} - \mathbf{A\hat{x}}]_j^2} \label{betaupdate} ,
\end{align}
where $\Sigma_{ii} = [\boldsymbol{\Sigma}]_{ii}$. The derivations of \eqref{alphaupdate} and \eqref{betaupdate} are given in the next section.




\subsection{Derivation of update equations for SD-RVM}

\label{subsec:derivation_SD-RVM}

To update the precisions we maximize the distribution $p(\mathbf{y},\boldsymbol{\gamma}, \boldsymbol{\beta}) = p(\mathbf{y}|\boldsymbol{\gamma}, \boldsymbol{\beta})p(\boldsymbol{\gamma})p(\boldsymbol{\beta})$ (obtained by marginalizing over $\mathbf{x}$), with respect to $\gamma_i$ and $\beta_j$, where we use the prior
\begin{align*}
p(\beta_j) = \mathrm{Gamma}(\beta_j | c+1,d),
\end{align*}
and $p(\gamma_i)$ is as in \eqref{alphaprior}. The log-likelihood of the parameters is
\begin{align}
\mathcal{L} = &\text{constant} - \frac{1}{2} \log \mathrm{det}(\mathbf{B}^{-1} + \mathbf{A} \boldsymbol{\Gamma}^{-1} \mathbf{A^\top}) \label{loglikelihood}\\
&- \frac{1}{2} \mathbf{y}^\top (\mathbf{B}^{-1} + \mathbf{A}\boldsymbol{\Gamma}^{-1} \mathbf{A}^\top)^{-1} \mathbf{y} \nonumber\\
& + \sum_{i=1}^n (a\log \gamma_i - b\gamma_i) + \sum_{j=1}^m (c\log \beta_j - d\beta_j) . \nonumber
\end{align}

We maximize $\mathcal{L}$ w.r.t. $\gamma_i$ by setting the derivative to zero. To simplify the derivative we use that
\begin{align}
\label{eq:xhat_identity}
\frac{\partial}{\partial \gamma_i} \left( \mathbf{y}^\top (\mathbf{B}^{-1} + \mathbf{A}\boldsymbol{\Gamma}^{-1} \mathbf{A}^\top)^{-1} \mathbf{y} \right) = \hat{x}_i^2 , 
\end{align}
and the determinant lemma \cite{harville08}
\begin{align}
\label{determinantlemma}
\mathrm{det}(\mathbf{B}^{-1} + \mathbf{A}\boldsymbol{\Gamma}^{-1} \mathbf{A}^\top) 
= \mathrm{det}(\boldsymbol{\Sigma}^{-1}) 
\mathrm{det}(\boldsymbol{\Gamma}^{-1}) \mathrm{det}(\mathbf{B}^{-1}) .
\end{align}
Using \eqref{eq:xhat_identity} and \eqref{determinantlemma} we find that $\mathcal{L}$ is maximized w.r.t. $\gamma_i$ when
\begin{align}
\label{derivativezero}
 - \frac{1}{2} \Sigma_{ii} + \frac{1}{2\gamma_i} + \frac{a}{\gamma_i} - b - \frac{1}{2} \hat{x}_i^2 = 0. 
\end{align}
Instead of solving for $\gamma_i$ (which would require solving a non-linear coupled equation since $\boldsymbol{\Sigma}$ and $\mathbf{\hat{x}}$ depend on $\gamma_i$) we approximate the equation as
\begin{align}
\label{alphanew1}
1 - \gamma_i \Sigma_{ii} + 2a - (\hat{x}_i^2 + 2b) \gamma_i^{new} = 0 . 
\end{align}
We solve \eqref{alphanew1} for $\gamma_i^{new}$ rather than \eqref{derivativezero} for $\gamma_i$ since it in practice often results in a better convergence \cite{Tipping1,Mackay}. The update equation then becomes
\begin{align*}
\gamma_{i}^{new} = \frac{1 - \gamma_i \Sigma_{ii} + 2a}{\hat{x}_i^2 + 2b} . 
\end{align*}
Setting $a = b = 0$ we obtain \eqref{alphaupdate}.

For the noise precisions we use that
\begin{align}
\label{eq:res_identity}
&\frac{\partial}{\partial \beta_j} \left[ \mathbf{y}^\top (\mathbf{B}^{-1} +\mathbf{A}\boldsymbol{\Gamma}^{-1} \mathbf{A}^\top)^{-1} \mathbf{y} \right]
= [\mathbf{y - A\hat{x}}]_j^2 .
\end{align}
We find that $\mathcal{L}$ is maximized w.r.t. $\beta_j$ when
\begin{align*}
- \frac{1}{2} \mathrm{tr}(\boldsymbol{\Sigma} \mathbf{A}_{j,:}^\top \mathbf{A}_{j,:}) + \frac{1}{2\beta_j} - \frac{1}{2} [\mathbf{y - A\hat{x}} ]_j^2 + \frac{c}{\beta_j} - d = 0 , 
\end{align*}
where $\mathbf{A}_{j,:}$ denotes the $j$'th row vector of $\mathbf{A}$. Rewriting the equation as
\begin{align*}
1 - \beta_j \mathbf{A}_{j,:} \boldsymbol{\Sigma} \mathbf{A}_{j,:}^\top + 2c - ([\mathbf{y - A\hat{x}}]_j^2 + 2d) \beta_j^{new} = 0 , 
\end{align*}
using that $\mathbf{A}_{j,:} \boldsymbol{\Sigma}\mathbf{A}_{j,:}^\top = [\mathbf{A}\boldsymbol{\Sigma}\mathbf{A}^\top]_{jj}$, we find that
\begin{align*}
\beta_j^{new} = \frac{1 - \beta_j [\mathbf{A}\boldsymbol{\Sigma}\mathbf{A}^\top]_{jj} + 2c}{[\mathbf{y - A\hat{x}}]_j^2 + 2d} . 
\end{align*}
Setting $c = d= 0$ we obtain \eqref{betaupdate}.

The derivations of \eqref{eq:xhat_identity} and \eqref{eq:res_identity} are given in Appendix~\ref{xhat_identity_derivation}.



\subsection{Analysis of sparsity}

\label{motivation}


Several approximations are made in the derivation of the iterative update equations. It is interesting to see how the approximations affect the sparsity of the solution. In this subsection, we show that the approximations make the SD-RVM equivalent to minimizing a non-symmetric sparsity promoting cost function.

To motivate that the standard RVM is sparsity promoting, one can use that the marginal distribution of $x_i$ is a student-t distribution. For a fixed $\beta$ (and $\mathbf{e = 0}$), the standard RVM is therefore an iterative method for solving (details can be found in \cite{Tipping1})
\begin{align*}
\min_{\mathbf{x}} \frac{\beta}{2} ||\mathbf{y - Ax}||_2^2 + \left(1+\frac{a}{2}\right) \sum_{i=1}^n   \log(x_i^2 + 2b) .
\end{align*}
The log-sum cost function can be used as a sparsity promoting cost function, making it plausible that the RVM promotes sparsity.

\begin{figure}[t]
\includegraphics[width = 1.0\columnwidth]{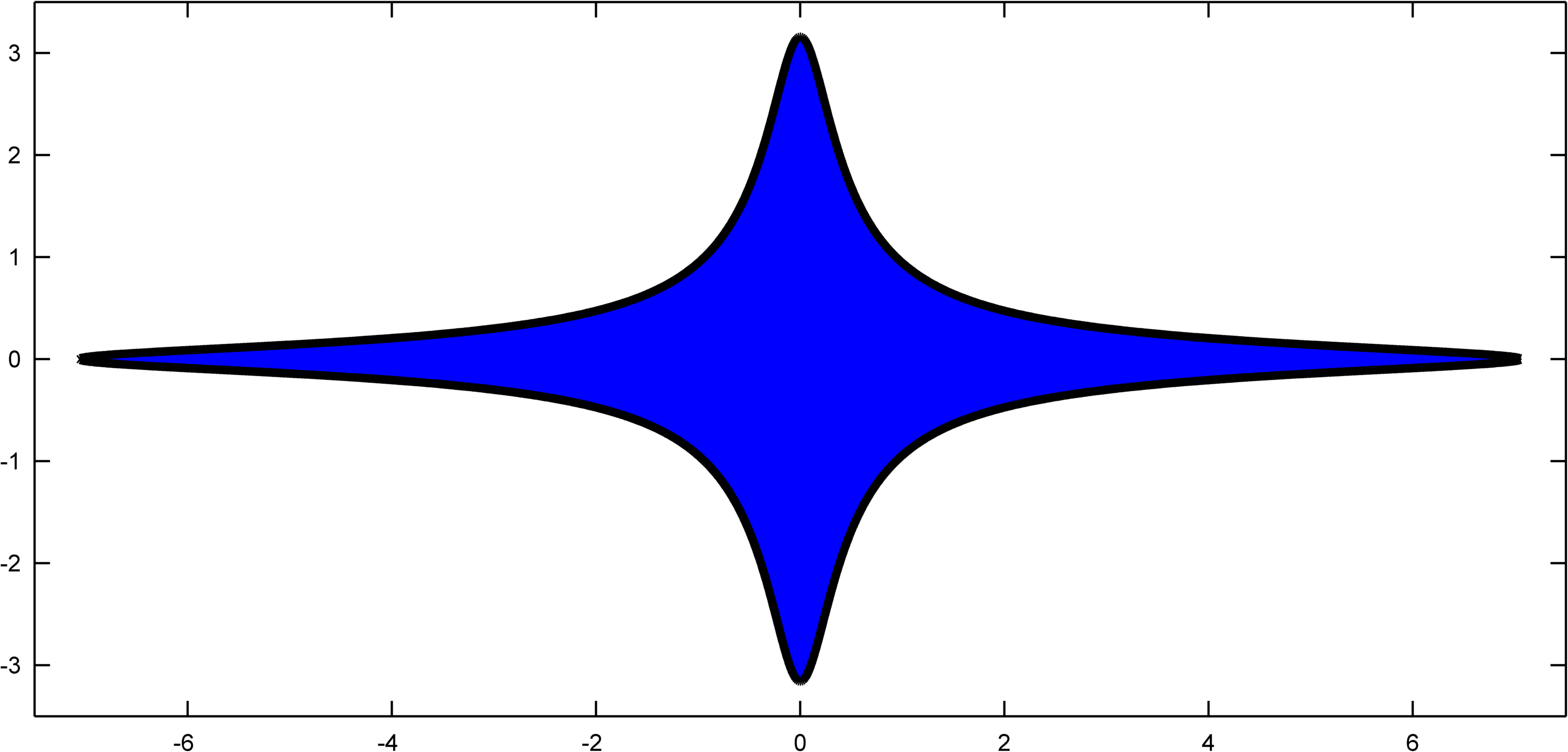}
\caption{The non-symmetric \emph{log-ball} $\log(x_1^2 + 0.02) + \log(x_2^2 + 0.1) \leq 0$. SD-RVM is equivalent to finding the smallest non-symmetric log-ball that intersects the linear subspace $\mathbf{Ax + \tilde{e} = y}$.}
\label{logball}
\end{figure}

For the SD-RVM, the precisions are updated by maximizing the marginal distribution $p(\mathbf{y},\boldsymbol{\gamma},\boldsymbol{\beta})$. The problem is equivalent to maximizing $\mathcal{L}$ in \eqref{loglikelihood}. We show approximations for relevant parts of the right hand side of $\mathcal{L}$ as follows
\begin{align}
&\log \mathrm{det} (\boldsymbol{\Sigma}^{-1}) \approx \log \mathrm{det} ((\boldsymbol{\Sigma}^{old})^{-1})  +  \nonumber \\ 
&  \sum_{i=1}^n \Sigma_{ii}^{old} (\gamma_i - \gamma_i^{old}) + \sum_{j=1}^m [\mathbf{A}\boldsymbol{\Sigma}^{old} \mathbf{A}^\top]_{jj} (\beta_j - \beta_j^{old}),\label{eq:approx_1}
\end{align}
where the approximation is up to first order in $\boldsymbol{\gamma}$ and $\boldsymbol{\beta}$. 
We rewrite the problem in variables $\mathbf{x}$ and $\mathbf{\tilde{e}}$ using that \cite{Rojas}
\begin{align}
\mathbf{y}^\top (\mathbf{A}\boldsymbol{\Gamma}^{-1} \mathbf{A}^\top + \mathbf{B}^{-1})^{-1} \mathbf{y} = &\min_{\mathbf{x,\tilde{e}}} \sum_{i=1}^n \gamma_i x_i^2 + \sum_{j=1}^m \beta_j \tilde{e}_j^2 , \label{eq:approx_2} \\ 
&\text{such that  } \mathbf{Ax + \tilde{e} = y} \nonumber
\end{align}
where now $\mathbf{\tilde{e} = e + n}$ as in \eqref{robustbeta}. Under the approximation \eqref{eq:approx_1} and the reformulation \eqref{eq:approx_2}, maximization of $\log p(\mathbf{y},\boldsymbol{\gamma},\boldsymbol{\beta})$ becomes equivalent to
\begin{align*}
&\min_{\gamma_i,\beta_j,\mathbf{x,\tilde{e}}} \sum_{i=1}^n \left[ (x_i^2 + \Sigma_{ii}^{old} + 2b)\gamma_i + (1+2a)\log(\gamma_i) \right] \\
&+ \sum_{j=1}^m \left[(e_j^2 + [\mathbf{A}\boldsymbol{\Sigma}^{old} \mathbf{A}^\top]_{jj} + 2d)\beta_j + (1+2c)\log(\beta_j) \right]  . \nonumber\\
&\text{such that  } \mathbf{Ax + \tilde{e} = y} \nonumber
\end{align*}
By minimizing the objective with respect to $\gamma_i$ and $\beta_j$, the problem reduces to
\begin{align}
&\min_{\mathbf{x,\tilde{e}}} \, \, (1+2a) \sum_{i=1}^n \log(x_i^2 + \Sigma_{ii}^{old} + 2b) \label{logmin}\\
&+ (1 + 2c) \sum_{j=1}^m \log(\tilde{e}_j^2 + [\mathbf{A}\boldsymbol{\Sigma}^{old}\mathbf{A}^\top]_{jj} + 2d) , \nonumber\\
&\text{such that  } \mathbf{Ax + \tilde{e} = y} \nonumber
\end{align}
where we have ignored additive constants. Because of the approximations, the constants $\Sigma_{ii}^{old}$ and $[\mathbf{A}\boldsymbol{\Sigma}^{old}\mathbf{A}^\top]_{jj}$ make the cost function non-symmetric in the components of $\mathbf{x}$ and $\mathbf{\tilde{e}}$. The SD-RVM is thus equivalent to minimizing a non-symmetric sparsity promoting cost function.
In a similar way it can be shown that the standard RVM and RB-RVM are also equivalent to non-symmetric cost functions under appropriate approximations.
A two-dimensional example using $\mathbf{x}=[x_1, \, x_2]^{\top}$ is shown in Fig.~\ref{logball}.

\subsection{Computational complexity}

\label{cputimes:theory}

In this section we take a non-rigorous approach for quantifying the computational complexity of SD-RVM. 
The complexity is computed per iteration, since the number of iterations depends on the stopping criterion used, and with the assumption of a naive implementation. Each iteration of SD-RVM requires $\mathcal{O}(n^3)$ flops to compute the matrix $\boldsymbol{\Sigma}$
using Gauss-Jordan elimination \cite{num_linalg}. 
Updating the precisions requires $\mathcal{O}(nm)$ flops since the residual $\mathbf{y - A \hat{x}}$ needs to be computed. 
Hence the computational complexity of SD-RVM is 
\begin{align*}
\mathcal{O}(\mathrm{max}(nm,n^3)) = \mathcal{O}(n \cdot \mathrm{max} (m,n^2)).
\end{align*}
A natural interest is the complexity of RB-RVM. Again with the assumption of a naive implementation,
each iteration of RB-RVM requires the inversion of a $(n+m) \times (n+m)$ matrix to compute $\boldsymbol{\Sigma}_{RB}$.
Updating the precisions requires $\mathcal{O}(nm)$ flops and hence the computational complexity of RB-RVM is
\begin{align*}
\mathcal{O}(\mathrm{max}(nm,(n+m)^3)) = \mathcal{O}((n+m)^3). 
\end{align*}
In Section \ref{cs_problem} we provide numerical evaluations to quantify
algorithm run time requirements that confirm that SD-RVM is typically faster than RB-RVM.

\section{SD-RVM with Block Structure}

\subsection{SD-RVM for known block structure}

\label{SD-RVM_blocks}

To describe a block sparse signal $\mathbf{x} \in \mathbb{R}^n$ with known block structure we partition $[n] = \{1,2,\dots,n\}$ into blocks as
\begin{align*}
[n] = I_1 \cup I_2 \cup \dots \cup I_p ,
\end{align*}
where $|I_i| = n_i$ and $I_i \cap I_j = \emptyset$ for $i \neq j$. The signal is block sparse when only a few blocks of the signal are non-zero. 
The component-wise SD-RVM generalizes to this scenario by requiring that the precisions are equal in each block, i.e. we choose the prior distribution for the components of block $I_i$ to be
\begin{align*}
\mathbf{x}_{I_i} \sim \mathcal{N}( \mathbf{0},\gamma_i^{-1} \mathbf{I}_{n_i}) . 
\end{align*}
where $\mathbf{x}_{I_i} \in \mathbb{R}^{n_i}$ denotes the vector consisting of the components of $\mathbf{x}$ with indices in $I_i$.

Similarly we can partition the components of the sparse noise $\mathbf{\tilde{e}} \in \mathbb{R}^m$ into blocks as
\begin{align*}
[m] = J_1 \cup J_2 \cup \dots \cup J_q ,
\end{align*}
where $|J_j| = m_j$, $J_j \cap J_i = \emptyset$ for $i \neq j$ and the block $J_j$ of $\mathbf{e}$ is given the prior distribution
\begin{align*}
\mathbf{\tilde{e}}_{J_j} \sim \mathcal{N}( \mathbf{0},\beta_j^{-1} \mathbf{I}_{m_j}) . 
\end{align*}
As before, the precisions are given gamma distributions \eqref{alphaprior} as priors, where now
\begin{align}
\label{betaprior}
p(\beta_j) = \mathrm{Gamma}(\beta_j | c+1,d) .
\end{align}
Using this model, we derive the update equations of precisions as below
\begin{align}
&\gamma_i^{new} = \frac{n_i - \gamma_i \, \mathrm{tr}(\boldsymbol{\Sigma}_{I_i}) + 2a}{||\mathbf{\hat{x}}_{I_i}||_2^2 + 2b}, \label{blockgamma_update}\\ 
&\beta_j^{new} = \frac{m_j - \beta_j \, \mathrm{tr}([\mathbf{A\boldsymbol{\Sigma} A^\top}]_{J_j}) + 2c}{||(\mathbf{y - A\hat{x}})_{J_j}||_2^2 + 2d}, \label{blockbetaupdate}
\end{align}
where $\boldsymbol{\Sigma}_{I_i}$ denotes the $n_i \times n_i$ submatrix of $\boldsymbol{\Sigma}$ formed by elements appropriately indexed by $I_i$. By setting $I_i = \{i\}$ and $J_j = \{ j \}$ we obtain the update equations for component-wise sparse signal and noise. We see that when $I_i = \{ i \}$ and $J_j = J = [m]$, then \eqref{blockbetaupdate} reduces to the update equations of the standard RVM since
\begin{align*}
n - \beta \, \mathrm{tr}(\mathbf{A\boldsymbol{\Sigma} A^\top}) = \sum_i \gamma_i \Sigma_{ii} . 
\end{align*}

The derivation of the update equations \eqref{blockgamma_update} and \eqref{blockbetaupdate} is found in Appendix~\ref{sec:derivation_SDRVM_knownblocks}.

\subsection{SD-RVM for unknown block structure}

\label{unknownblocks}

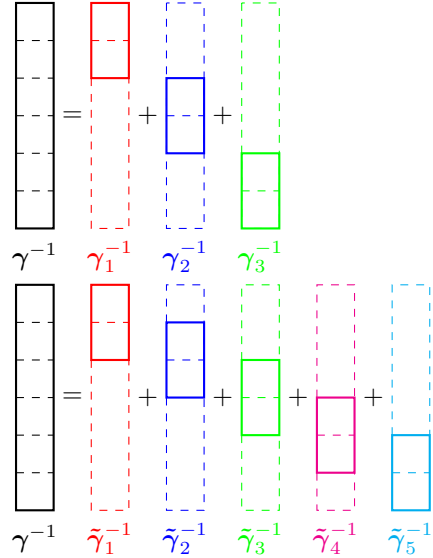
\begin{figure}
\begin{center}
\begin{tikzpicture}[scale = 0.5]
\draw[thick,color=black] (0,0) rectangle (1,6);
\draw[dashed,color=black] (0,1) -- (1,1);
\draw[dashed,color=black] (0,2) -- (1,2);
\draw[dashed,color=black] (0,3) -- (1,3);
\draw[dashed,color=black] (0,4) -- (1,4);
\draw[dashed,color=black] (0,5) -- (1,5);
\node at (0.5,-0.75) {$\boldsymbol{\gamma}^{-1}$};
\node at (1.5,3) {$=$};
\draw[dashed,color=red] (2,4) -- (2,0) -- (3,0) -- (3,4);
\draw[thick,color=red] (2,6) rectangle (3,4);
\draw[dashed,color=red] (2,5) -- (3,5);
\node[color=red] at (2.5,-0.75) {$\boldsymbol{\gamma}_1^{-1}$};
\node at (3.5,3) {$+$};
\draw[dashed,color=blue] (4,6) rectangle (5,0);
\draw[thick,color=blue] (4,4) rectangle (5,2);
\draw[dashed,color=blue] (4,3) -- (5,3);
\node[color=blue] at (4.5,-0.75) {$\boldsymbol{\gamma}_2^{-1}$};
\node at (5.5,3) {$+$};
\draw[dashed,color=green] (6,6) rectangle (7,0);
\draw[thick,color=green] (6,2) rectangle (7,0);
\node[color=green] at (6.5,-0.75) {$\boldsymbol{\gamma}_3^{-1}$};
\draw[dashed,color=green] (6,1) -- (7,1);
\draw[thick,color=black] (0,0-7.5) rectangle (1,6-7.5);
\draw[dashed,color=black] (0,1-7.5) -- (1,1-7.5);
\draw[dashed,color=black] (0,2-7.5) -- (1,2-7.5);
\draw[dashed,color=black] (0,3-7.5) -- (1,3-7.5);
\draw[dashed,color=black] (0,4-7.5) -- (1,4-7.5);
\draw[dashed,color=black] (0,5-7.5) -- (1,5-7.5);
\node at (0.5,-0.75-7.5) {$\boldsymbol{\gamma}^{-1}$};
\node at (1.5,3-7.5) {$=$};
\draw[dashed,color=red] (2,4-7.5) -- (2,0-7.5) -- (3,0-7.5) -- (3,4-7.5);
\draw[thick,color=red] (2,6-7.5) rectangle (3,4-7.5);
\draw[dashed,color=red] (2,5-7.5) -- (3,5-7.5);
\node[color=red] at (2.5,-0.75-7.5) {$\boldsymbol{\tilde{\gamma}}_1^{-1}$};
\node at (3.5,3-7.5) {$+$};
\draw[dashed,color=blue] (4,5-7.5) -- (4,6-7.5) -- (5,6-7.5) -- (5,4-7.5);
\draw[thick,color=blue] (4,5-7.5) rectangle (5,3-7.5);
\draw[dashed,color=blue] (4,3-7.5) -- (4,0-7.5) -- (5,0-7.5) -- (5,3-7.5);
\node[color=blue] at (4.5,-0.75-7.5) {$\boldsymbol{\tilde{\gamma}}_2^{-1}$};
\draw[dashed,color=blue] (4,4-7.5) -- (5,4-7.5);
\node at (5.5,3-7.5) {$+$};
\draw[dashed,color=green] (6,4-7.5) -- (6,6-7.5) -- (7,6-7.5) -- (7,4-7.5);
\draw[thick,color=green] (6,4-7.5) rectangle (7,2-7.5);
\draw[dashed,color=green] (6,2-7.5) -- (6,0-7.5) -- (7,0-7.5) -- (7,2-7.5);
\node[color=green] at (6.5,-0.75-7.5) {$\boldsymbol{\tilde{\gamma}}_3^{-1}$};
\draw[dashed,color=green] (6,3-7.5) -- (7,3-7.5);
\node at (7.5,3-7.5) {$+$};
\draw[dashed,color=magenta] (8,3-7.5) -- (8,6-7.5) -- (9,6-7.5) -- (9,3-7.5);
\draw[thick,color=magenta] (8,3-7.5) rectangle (9,1-7.5);
\draw[dashed,color=magenta] (8,1-7.5) -- (8,0-7.5) -- (9,0-7.5) -- (9,1-7.5);
\node[color=magenta] at (8.5,-0.75-7.5) {$\boldsymbol{\tilde{\gamma}}_4^{-1}$};
\draw[dashed,color=magenta] (8,2-7.5) -- (9,2-7.5);
\node at (9.5,3-7.5) {$+$};
\draw[dashed,color=cyan] (10,2-7.5) -- (10,6-7.5) -- (11,6-7.5) -- (11,2-7.5);
\draw[thick,color=cyan] (10,2-7.5) rectangle (11,0-7.5);
\node[color=cyan] at (10.5,-0.75-7.5) {$\boldsymbol{\tilde{\gamma}}_5^{-1}$};
\draw[dashed,color=cyan] (10,1-7.5) -- (11,1-7.5);
\end{tikzpicture}
\end{center}
\caption{Illustration of non-overlapping and overlapping block parameterizations.}
\label{blocks}
\end{figure}

In some situations the signal can have an unknown block structure, i.e. the signal is block sparse, but the dimensions and positions of the blocks are unknown. This scenario can be handled by treating the signal as a superposition of block sparse signals \cite{zhang} (see illustration in Figure~\ref{blocks}). This approach also describes the scenario \eqref{eq:sparsenoise} when $\mathbf{e}$ is component wise sparse and $\mathbf{n}$ is dense (e.g. Gaussian). The precision of each component is then a combination of the precisions of the blocks to which the component belongs. Let $\gamma_i$ be the precision of the component $x_i$ and $\tilde{\gamma}_k$ be the precision of block $I_k$. We model the signal as
\begin{align}
&x_i \sim \mathcal{N}( 0,\gamma_i^{-1}) , \nonumber \\
&\gamma_i^{-1} = \sum_{k, i \in I_k} \tilde{\gamma}_k^{-1} . \label{alphasum}
\end{align}
We model the noise in a similar way with precisions $\beta_j$ for component $j$ and precisions $\tilde{\beta}_l$ for the block with support $J_l$. To promote sparsity, the precisions of the underlying blocks are given gamma distributions as priors
\begin{align*}
&\tilde{\gamma}_k \sim \mathrm{Gamma}(\tilde{\gamma}_k | a+1,b) ,\\
&\tilde{\beta}_l \sim \mathrm{Gamma}(\tilde{\beta}_l | c+1,d) .
\end{align*}
In each iteration we update the underlying precisions $\tilde{\gamma}_k$. The componentwise precisions are then updated using \eqref{alphasum}. With this model, the update equations for the precisions become
\begin{align}
&\tilde{\gamma}_k^{new} = \frac{\frac{1}{\tilde{\gamma}_k}\mathrm{tr}(\boldsymbol{\Gamma}_k)  - \frac{1}{\tilde{\gamma}_k} \mathrm{tr} (\boldsymbol{\Gamma}_k \boldsymbol{\Sigma} \boldsymbol{\Gamma}_k) + 2a}{\frac{1}{\tilde{\gamma}_k^2} || \boldsymbol{\Gamma}_i \mathbf{\hat{x}}||_2^2 + 2b} ,\label{gamma_unknown_update}\\
&\tilde{\beta}_l^{new} = \frac{\frac{1}{\tilde{\beta}_l} \mathrm{tr}(\mathbf{B}_l) - \frac{1}{\tilde{\beta}_l} \mathrm{tr}(\mathbf{B}_l \mathbf{A}^\top\boldsymbol{\Sigma}\mathbf{A}\mathbf{B}_l) + 2c}{\frac{1}{\tilde{\beta}_l^2} || \mathbf{B}_l (\mathbf{y - A\hat{x}})||_2^2 + 2d} , \label{beta_unknown_update}
\end{align}
where $\boldsymbol{\Gamma}_k$ is the diagonal matrix with $[\boldsymbol{\Gamma}_k]_{ii} = \gamma_i$ if $i \in I_k$ and $[\boldsymbol{\Gamma}_k]_{ii} = 0$ otherwise. We denote the corresponding matrix for $\beta_l$ by $\mathbf{B}_l$. The componentwise precisions are updated using \eqref{alphasum} and similar for $\beta_j$.


We see that when the underlying blocks are disjoint, then $\gamma_i = \tilde{\gamma}_k$ for all $i \in I_k$ and $\beta_j = \tilde{\beta}_l$ for all $j \in J_l$. The update equations then reduce to the update equations \eqref{blockbetaupdate} for the block sparse model with known block structure.

\subsubsection{Sparse and dense noise}

\label{sec:sparse_and_dense}

In the model \eqref{eq:sparsenoise} where $\mathbf{x}$ and $\mathbf{e}$ are componentwise sparse and $\mathbf{n}$ is dense, then
\begin{align}
\label{sparsegaussian_parametrization}
I_i = \{ i \}, \,\, J_j = \{ j \} , \,\, J_{m+1} = [m] , 
\end{align}
where $i=1,2,\dots , n$ and $j=1,2,\dots, m$. In this scenario the support set of the sparse and dense noise is overlapping, so the update equations for the precisions become
\begin{align*}
&\gamma_i^{new} = \frac{1 - \Sigma_{ii}\gamma_i + 2a}{\hat{x}_i^2 + 2b} ,\\
&\tilde{\beta}_j^{new} = \frac{1 - \frac{\beta_j}{\tilde{\beta}_j} [\mathbf{A}^\top \boldsymbol{\Sigma} \mathbf{A}]_{jj} + 2c}{\beta_j [\mathbf{y - A\hat{x}} ]_j^2 + 2d} ,\, \, j =1,2,\dots ,m ,\\
&\tilde{\beta}_{m+1}^{new} = \frac{\sum_{j=1}^m \beta_j - \frac{1}{\tilde{\beta}_{m+1}} \sum_{j=1}^m \beta_j^2 [\mathbf{A}^\top \boldsymbol{\Sigma} \mathbf{A}]_{jj} + 2c}{ \sum_{j=1}^m \beta_j^2 [\mathbf{y - A\hat{x}} ]_j^2 + 2d},\\
&\beta_j = (\tilde{\beta}_j^{-1} + \beta_{m+1}^{-1})^{-1} . 
\end{align*}
We will use these update equations in the simulations where the signal is component-wise sparse and 
the noise is a sum of (component-wise) sparse and dense noise. It turns out that this method is slightly better than the SD-RVM in section~\ref{subsec:SD-RVM}.


\section{Simulation experiments}

\label{sec:simulations}

In this section we evaluate the performance of the SD-RVM using several scenarios -- for simulated and real signals. 
For simulated signals, we considered the sparse and block sparse recovery problem in compressed sensing.
Then for real signals, we considered prediction of house prices using the Boston housing dataset \cite{boston} and denoising of images contaminated by \emph{salt and pepper} noise. In the simulations we used the cvx toolbox \cite{cvx} to implement JP.

\subsection{Compressed sensing}

\label{cs_problem}

\begin{figure}[t]
\begin{center}
\includegraphics[width=0.9\linewidth]{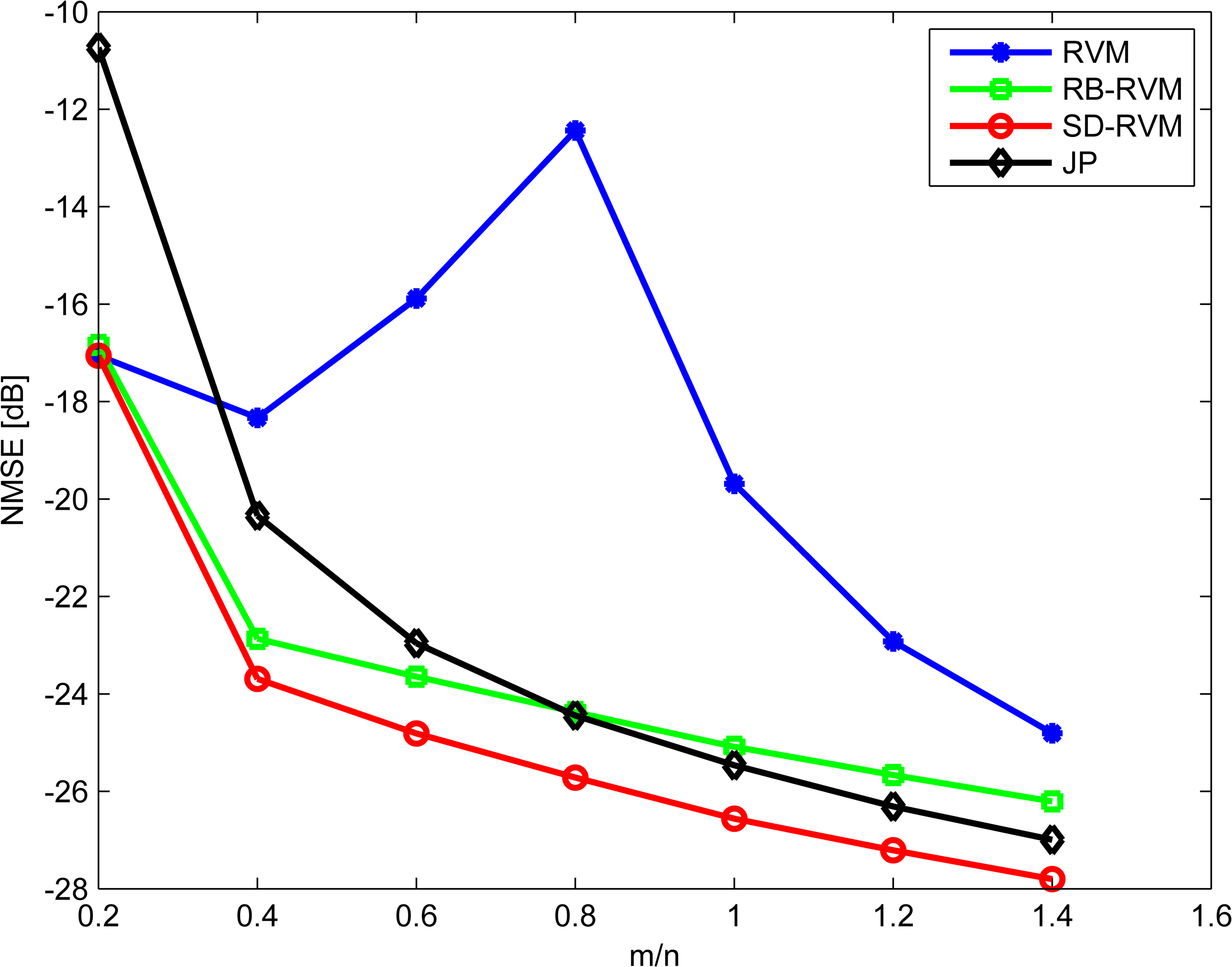}
\end{center}
\vspace{-0.5cm}
\caption{NMSE vs. $m/n$ for outlier free measurements.}
\label{CS1}
\end{figure}

\begin{figure}[t]
\begin{center}
\includegraphics[width=0.9\linewidth]{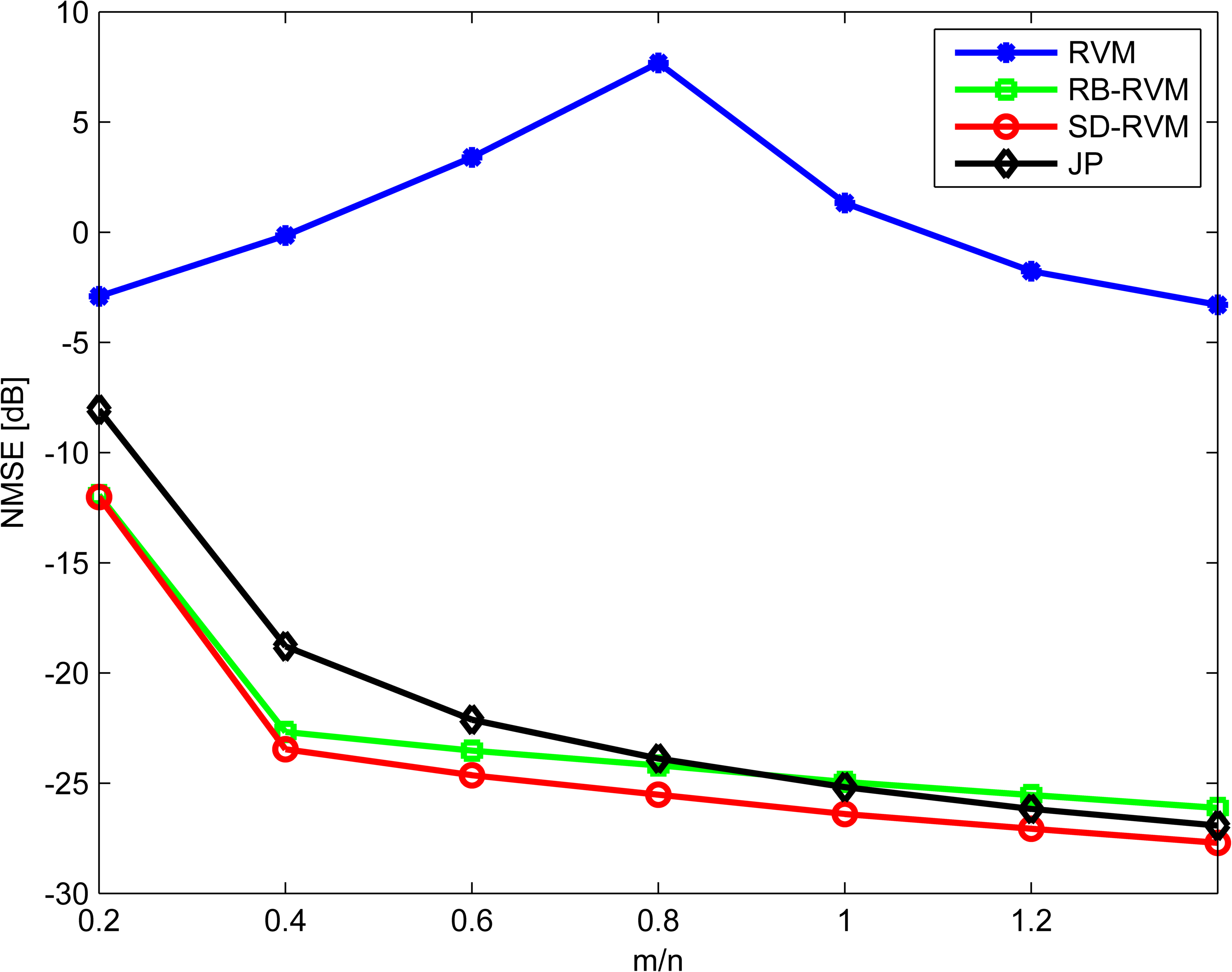}
\end{center}
\vspace{-0.5cm}
\caption{NMSE vs. $m/n$ for $5\%$ outliers contaminated measurements.}
\label{CS2}
\end{figure}

\begin{figure}[t]
\begin{center}
\includegraphics[width=0.9\linewidth]{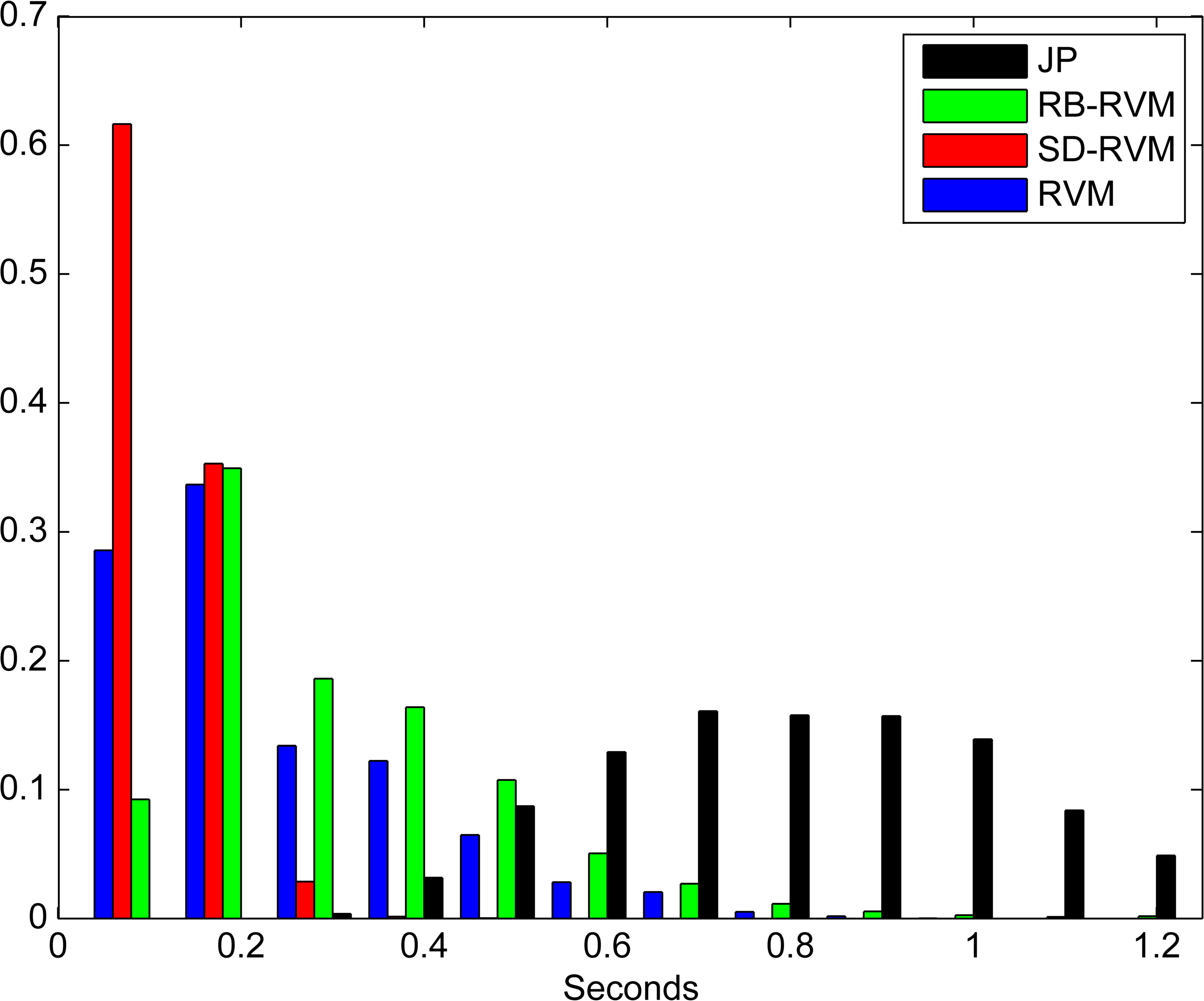}
\end{center}
\caption{Histogram of cputimes for the compressed sensing problem. Time is in seconds.}
\label{fig:cputimes}
\end{figure}

The recovery problem in compressed sensing consists of estimating a sparse vector $\mathbf{x} \in \mathbb{R}^n$ in \eqref{eq:sparsenoise} from $m$ linear measurements, where $m \ll n$. To evaluate the performance of the algorithms, we generated measurement matrices $\mathbf{A} \in \mathbb{R}^{m \times n}$ by drawing their components from a $\mathcal{N}(0,1)$ distribution and scaling their column vectors to unit norm. We selected the positions of the active components of $\mathbf{x}$ and $\mathbf{e}$ uniformly at random and draw their values from $\mathcal{N}(0,1)$. In the simulation we draw the additive noise $\mathbf{n}$ from $\mathcal{N}(\mathbf{0},\sigma_n^2 \mathbf{I}_m)$. We compared JP, the standard RVM, RB-RVM and SD-RVM. For JP \eqref{jp} we assumed $\sigma_n$ to be known and set $\epsilon = \sigma_n\sqrt{m + 2\sqrt{2m}}$ as proposed in \cite{Candes2}.

In the simulations we varied the measurement rate $m/n$ (ratio of the number of measurements and the signal dimension) for measurements without outliers and with $5\%$ outliers. We chose $n = 100$ and fixed the signal-to-dense-noise-ratio (SDNR)
\begin{align*}
\text{SDNR} = E[||\mathbf{Ax}||_2^2] / E[||\mathbf{n}||_2^2] = ||\mathbf{x}||_0 / (m \sigma_n^2) ,
\end{align*}
to $20$ dB. By generating $100$ measurement matrices and $100$ vectors $\mathbf{x}$ and $\mathbf{e}$ for each matrix we numerically evaluated the Normalized Mean Square Error (NMSE)
\begin{align*}
\text{NMSE} = E[|| \mathbf{x - \hat{x}}||_2^2] / E[||\mathbf{x}||_2^2] .
\end{align*}
The results are shown in Figure \ref{CS1} and Figure \ref{CS2}. We found that SD-RVM outperformed the other methods. The improvement of SD-RVM over RB-RVM was $1$ to $1.5$ dB for $m/n > 0.5$, with and without outliers. Compared to JP, the improvement of SD-RVM was $3$ to $3.7$ without outlier noise and $1$ to $4$ dB with outlier noise when $m/n > 0.5$. The poor performance of RVM is due to sensitivity to the regularization parameters. The performance of RVM improves greatly when the regularization are optimally tuned, however, the optimal values varies with SNR and measurement dimensions. The experiments show that the performance of SD-RVM does not degrade in the absence of sparse noise.

For each realization of the problem we measured the runtime (cpu time) of each algorithm. The histogram of the runtimes is shown in figure~\ref{fig:cputimes}. We found that the runtimes of the RVM algotithms (the standard RVM, RB-RVM and SD-RVM) were shorter than the runtime of JP and the runtimes of JP were spread over a larger range. Of the RVM algorithms, SD-RVM had the highest concentration of low runtime ($\leq 0.2$ seconds), while the runtimes of the standard RVM and RB-RVM was more concentrated around $0.2$ seconds. The histogram in figure~\ref{fig:cputimes} has been truncated to only show percentage for the visible values .

\subsection{Block sparse signals}

\begin{figure}[t]
\begin{center}
\includegraphics[width=0.9\linewidth]{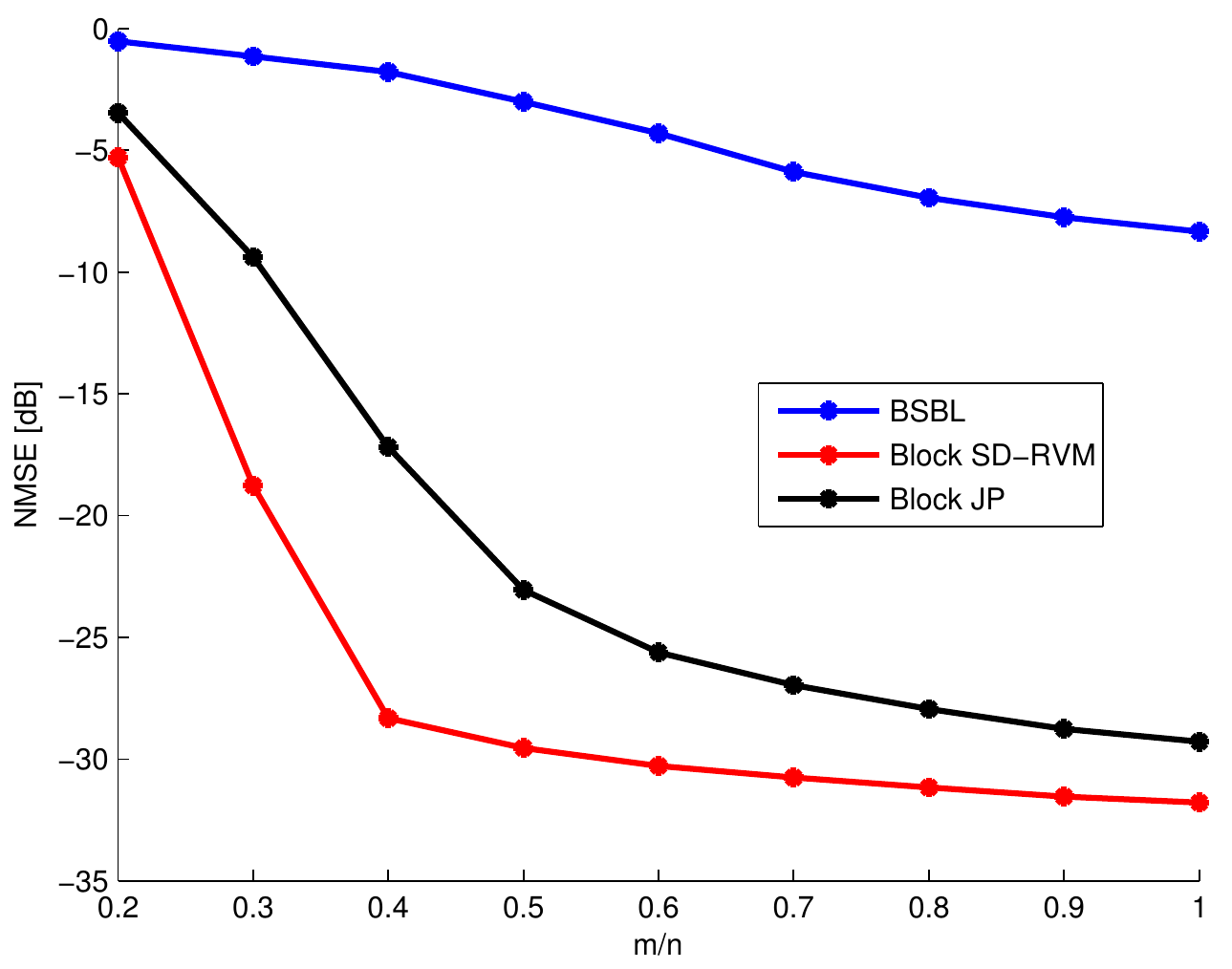}
\end{center}
\vspace{-0.5cm}
\caption{NMSE vs. $m/n$ for signals with known block structure and $5\%$ outliers in measurements.}
\label{blockCS1}
\end{figure}

\begin{figure}[t]
\begin{center}
\includegraphics[width=0.9\linewidth]{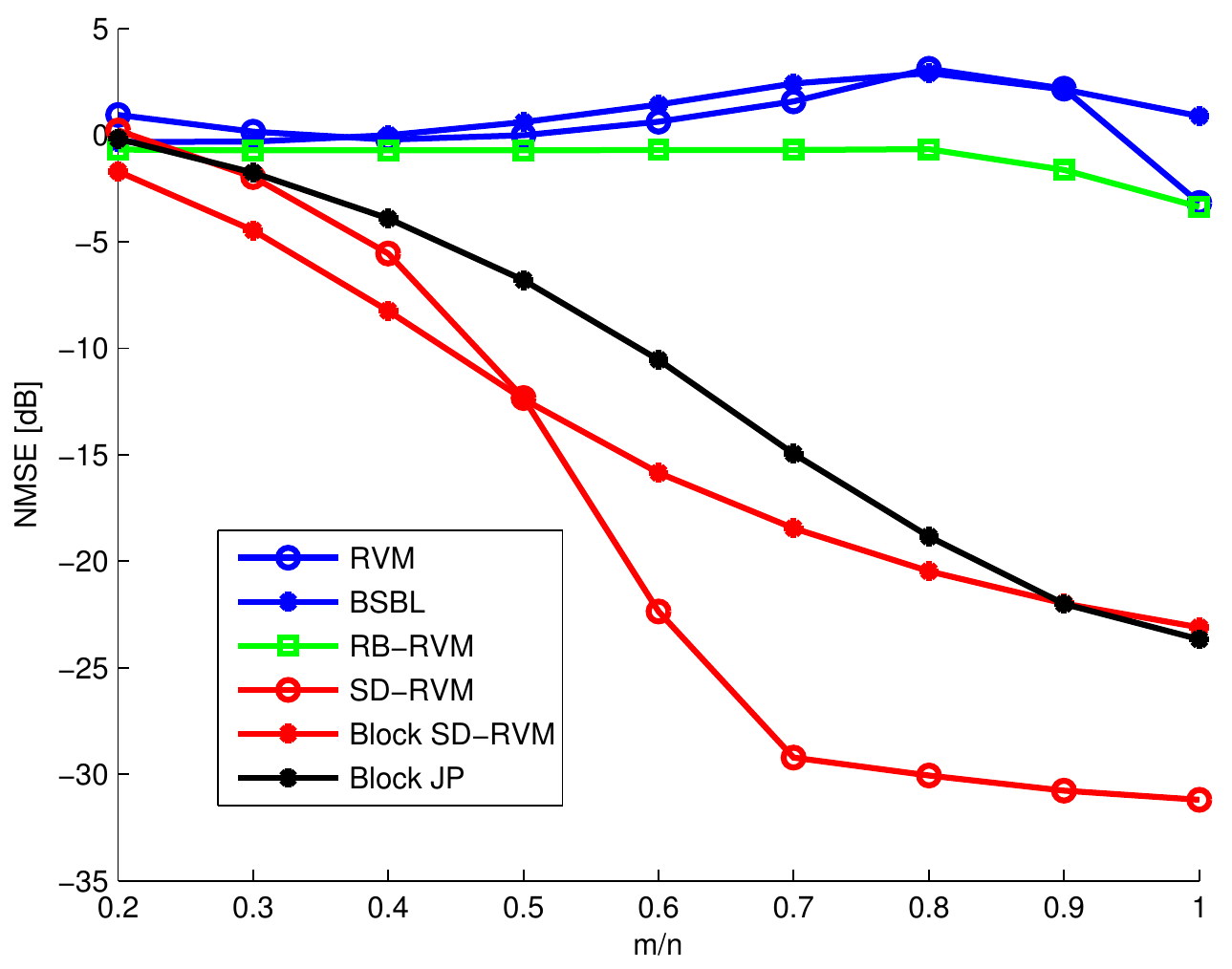}
\end{center}
\vspace{-0.5cm}
\caption{NMSE vs. $m/n$ for signals with unknown block structure and $5\%$ outliers in measurements.}
\label{blockCS2}
\end{figure}

The recovery problem in compressed sensing can be generalized to block sparse signals and noise \cite{modelCS}. For block sparse signals, the signal components are partitioned into blocks of which only a few blocks are non-zero. Sparse Bayesian learning (SBL) extended to the block sparse signal case is often referred to as block SBL (BSBL) \cite{zhang,zhang2}.  
The problem of unknown block structure can be solved by overparametrizing the blocks \cite{zhang}. In BSBL \cite{zhang}, the signal is modelled as
\begin{align*}
\mathbf{Ax} = \sum_{I \subset [n]} \mathbf{A}_I \mathbf{x}_I,
\end{align*}
i.e. the measured signal is modelled as a sum of signals where each signal represents a block of the original signal. The resulting problem can then be solved using BSBL for known block structure. When the minimum block size $K_{min}$ is known, the summation can be restricted to subsets of size $|I| = K_{min}$ \cite{zhang}.

The SD-RVM can be extended to the block sparse case using the methods developed in Section~\ref{SD-RVM_blocks} and Section~\ref{unknownblocks}. Justice Pursuit can be extended to the block sparse case in a similar way as BSBL by setting
\begin{align}
\mathbf{\hat{x},\hat{e}} = &\arg \min_{\mathbf{x,e}} \sum_i ||\mathbf{x}_{I_i}||_2 + \sum_j ||\mathbf{e}_{J_j}||_2, \label{jp_unknownblocks}\\
&\text{ such that } ||\mathbf{y - Ax - e}||_2 \leq \epsilon \nonumber
\end{align}
where the sum runs over all blocks (non-overlapping or overlapping) and as before we assume the noise variance to be known and set $\epsilon = \sigma_n \sqrt{m + \sqrt{8m}}$. For unknown block structure we also compared with component sparse methods RVM, RB-RVM and SD-RVM.

To numerically evaluate the performance of the block sparse algorithms we varied the measurement rate $m/n$ for measurements with $5\%$ sparse noise. We set the signal dimension to $n = 100$ and fixed the SDNR to $20$ dB. We divided the signal $\mathbf{x}$ into $20$ blocks of equal size of which $3$ blocks were non-zero. The sparse noise consisted of blocks with $5$ components in each block. In the sparse noise, $5\%$ of the blocks were active. For known block structure, the blocks were choosen uniformly at random from a set of predefined and non-overlapping blocks while for unknown block structure, the first component of each block was choosen uniformly at random, making it possible for the blocks to overlap. The active components of the signal and the sparse noise were drawn from $\mathcal{N}(0,1)$. By generating $50$ measurement matrices $\mathbf{A}$ and $50$ signals $\mathbf{x}$ and sparse noises $\mathbf{e}$ for each matrix we numerically evaluated the NMSE.


For known block structure we found that block SD-RVM outperformed the other methods. The NMSE of the block sparse SD-RVM was lower than the NMSE of block JP by more than $10$ dB for $m/n = 0.3,0.4$ and from $2$ to $6$ dB lower for $m/n \geq 0.5$.
The results are presented in Figure~\ref{blockCS1}.

For unknown block structure we found that for $m/n < 0.5$, SD-RVM for unknown block structure gave best performance while for $m/n \geq 0.5$, the usual component sparse SD-RVM gave the best perfomance. The NMSE of JP for unknown block structure was about $5$ dB larger than the NMSE of block SD-RVM for $0.4 \leq m/n \leq 0.6$, while for $m/n \geq 0.9$ block JP gave a better NMSE than block SD-RVM.
As expected, RVM and BSBL gave poor performance since they are not developed to handle measurements with sparse noise.
The results are shown in figure~\ref{blockCS2}.

\subsection{House price prediction}

One real world problem is the prediction of house prices. To test the algorithms on real data, we used the Boston housing dataset \cite{boston}. The dataset consists of $506$ house prices in suburbs of Boston and $13$ parameters (air quality, accessibility, pupil-to-teacher ratio, etc.) for each house. The problem is to predict the median house price for part of the dataset (test data) using the complement dataset (training data) to learn regression parameters. We model the house prices as
\begin{align*}
p_i = \mathbf{w}_i^\top \mathbf{x} + n_i + e_i, 
\end{align*}
where $p_i$ is the price of house $i$, $\mathbf{w}_i \in \mathbb{R}^{13}$ contains the parameters of house $i$, $\mathbf{x}\in\mathbb{R}^{13}$ is the regression vector, $n_i$ is (Gaussian) noise and $e_i$ is a (possible) outlier. Very expensive or inexpensive houses can treated as outliers.
The goal is to estimate the median house price for the test set. We find the median by estimating the regression parameters and setting
\begin{align*}
\hat{m} = \mathrm{median}(\mathbf{W}^\top \mathbf{\hat{x}}),
\end{align*}
where $\mathbf{W}$ contains the parameters of the houses in the test set. It is believed that only a few parameters are important to the average customer, $\mathbf{x}$ can therefore be modelled as a sparse vector.

We used a fraction $\rho$ of the dataset as training data and the rest as test set. By choosing the training set uniformly at random we evaluated the mean absolute error of the predicted median and mean cputime (in seconds) over $1000$ realizations.

We found that SD-RVM gave $10\%$ to $5\%$ lower mean error than that of RB-RVM and the mean error of RB-RVM and SD-RVM was about $70\%$ lower than the error of the RVM (see Table~\ref{tab:boston}). The cputime of SD-RVM was $16\%$ to $25\%$ of the cputime of RB-RVM.

\begin{table}
\small
\begin{center}
\caption{Prediction of median houseprice using the Boston Housing dataset. Mean error and mean cputime (in seconds) for different fractions, $\rho$, of the dataset used as training set.}
\renewcommand{\arraystretch}{1.2}
\begin{tabular}{|l| p{0.6cm} p{1.1cm} | p{0.6cm} p{1.1cm} | p{0.6cm} p{1.1cm} |}
\hline
&  \multicolumn{2}{|c|}{\textbf{RVM}} &  \multicolumn{2}{|c|}{\textbf{RB-RVM}} &  \multicolumn{2}{|c|}{\textbf{SD-RVM}}\\
\hline \label{tab:boston}
$\rho$ & \textbf{Error} & \textbf{Cputime} & \textbf{Error} & \textbf{Cputime} & \textbf{Error} & \textbf{Cputime}\\
\hline
0.3 & 1.24 & 0.18 & 0.43 & 0.60 & 0.38 & 0.15\\
0.4 & 1.26 & 0.29 & 0.39 & 1.25 & 0.35 & 0.25\\
0.5 & 1.27 & 0.42 & 0.39 & 2.20 & 0.36 & 0.38\\
0.6 & 1.28 & 0.60 & 0.41 & 3.28 & 0.37 & 0.53\\
0.7 & 1.28 & 0.92 & 0.45 & 5.27 & 0.43 & 0.80\\
\hline
\end{tabular}
\end{center}
\vspace*{-0.3cm}
\end{table}

\subsection{Image denoising}

\label{imagedenoising}


\begin{figure}[t]
\begin{center}
\includegraphics[width=0.9\linewidth]{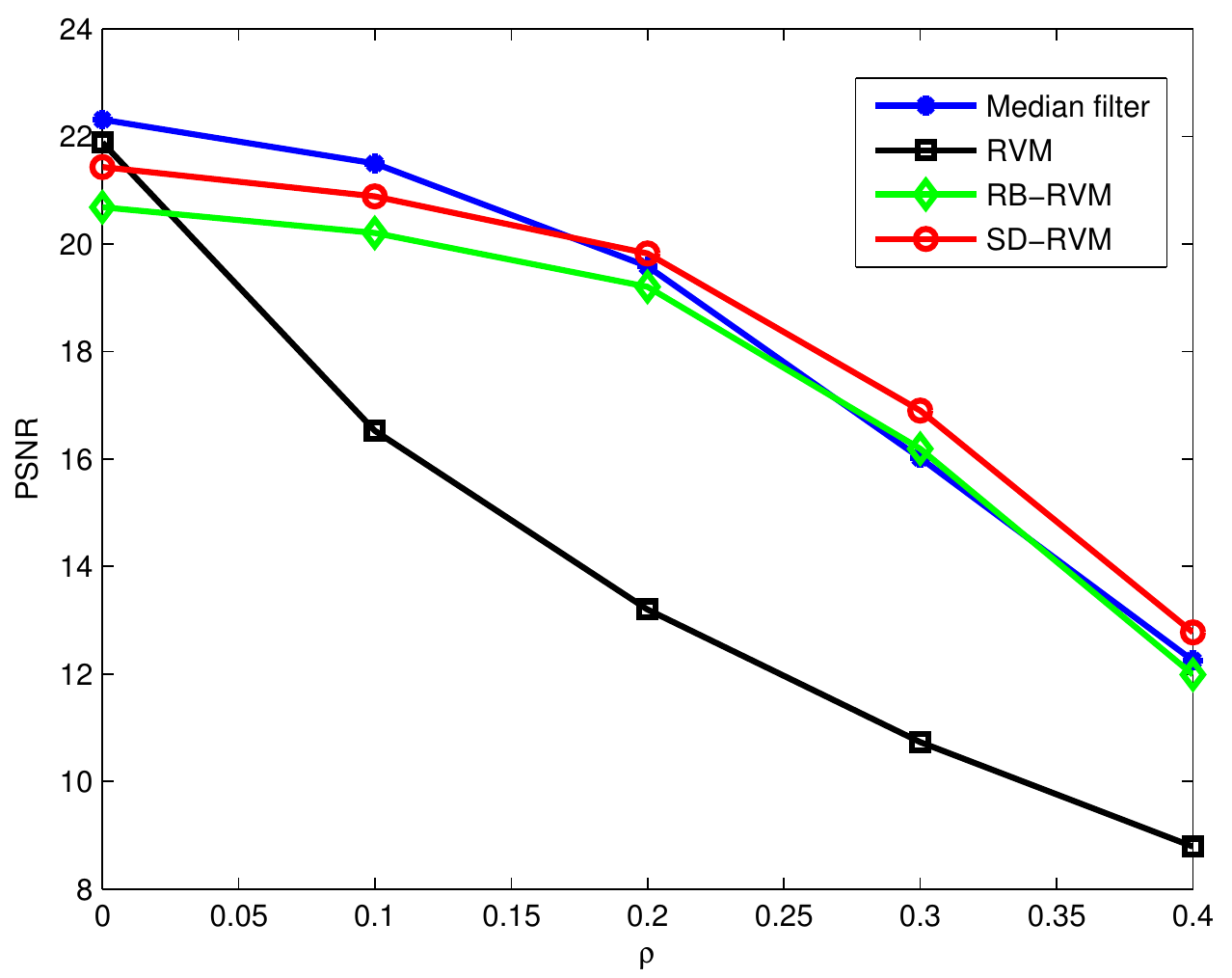}
\label{fig:error_rho}
\caption{PSNR vs. percentage of salt and pepper noise ($\rho$) averaged over $7$ images with 10 noise realizations for each image for each value of $\rho$.}
\label{fig:image1}
\end{center}
\end{figure}

\begin{figure*}[t]
\begin{center}
\includegraphics[width=0.9\textwidth]{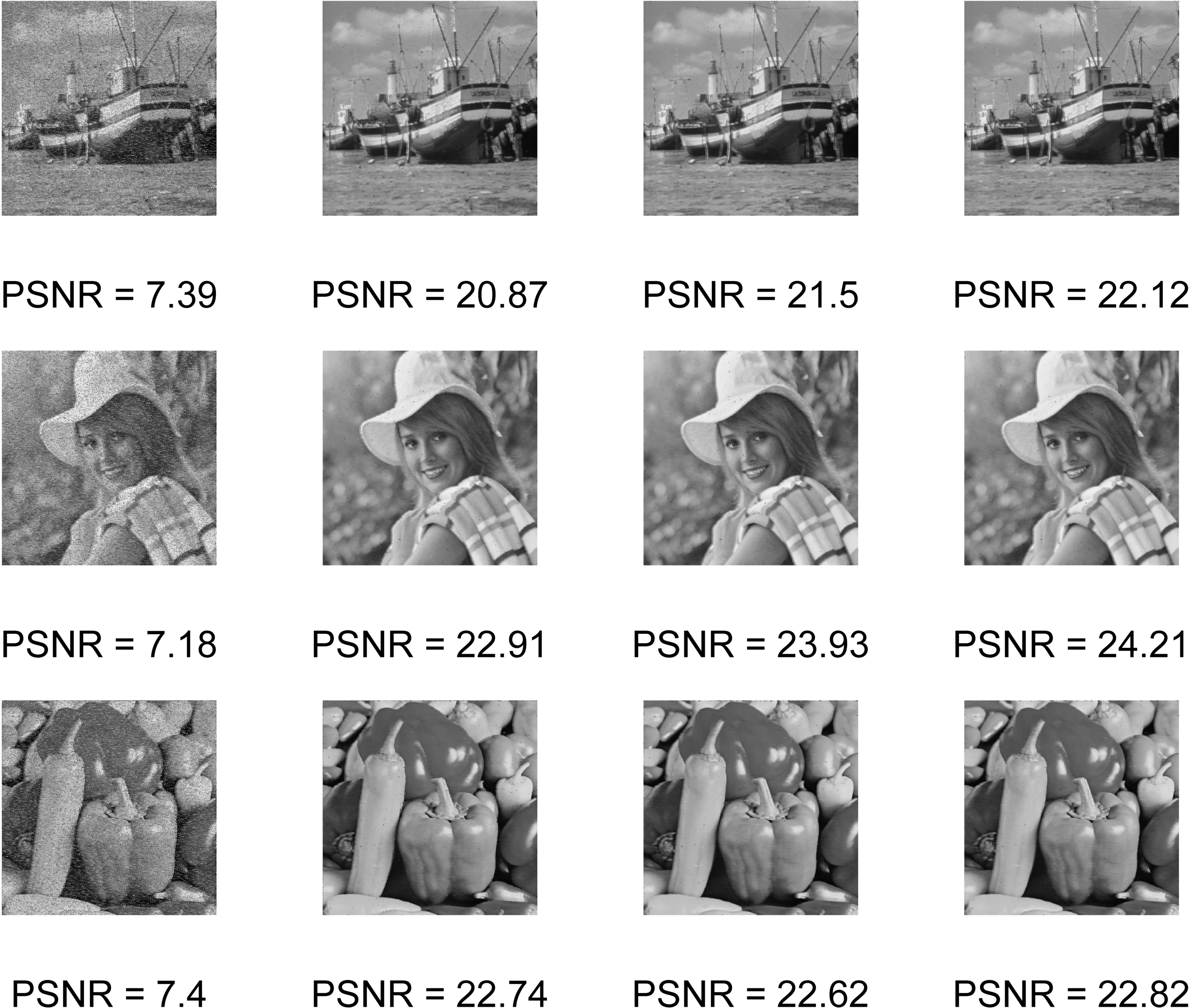}
\label{fig:image_test}
\caption{One realization of salt and pepper noise denoising for $\rho = 0.2$. Columns from left to right: Noisy image, median filter, RB-RVM and SD-RVM. Peak Signal to Noise Ratio (PSNR) has been rounded to two decimals.}
\label{fig:image_test1}
\end{center}
\end{figure*}

A grayscale image (represented in double-precision) can be modelled as an array of numbers in the interval from $0$ to $1$. Common sources of noise in images are electronic noise in sensors and bit quantization errors. Salt and Pepper \cite{Mitra2} noise makes some pixles black ($0$) or white ($1$). To test the algorithms we added $\rho$ percent of salt and pepper noise in $7$ different images (Boat, Baboon, Barbara, Elaine, House, Lena and Peppers) and denoised them using the median filter, the RVM, the RB-RVM and the SD-RVM. The pixels were set to either black or white with equal probability.

The median filter estimates the value of each pixel by the median in a $3 \times 3$ square patch. For the RVM, RB-RVM and SD-RVM, the value of a pixel was estimated by forming a $5 \times 5$ square patch around the pixel. In the patch, the pixels were modeled as \cite{Takeda}
\begin{align*}
y_i = \beta_0 + \boldsymbol{\beta}_1^\top (\mathbf{x - x}_i)  + \boldsymbol{\beta}_2^\top \mathrm{vech}( (\mathbf{x - x}_i)(\mathbf{x - x}_i)^\top) + n_i , 
\end{align*}
where $n_i$ is noise, $\mathbf{x}$ is the position of the central pixel, $\mathbf{x}_i$ is the position of pixel $i$, $i = 1,2,\dots, 25$ and $\mathrm{vech}$ is the \emph{half-vectorization operator} \cite{Takeda}, i.e.
\begin{align*}
\mathrm{vech} \left( \left( \begin{array}{cc}
a & b\\
b & c
\end{array} \right) \right) = \left( \begin{array}{c}
a\\b\\c
\end{array} \right) . 
\end{align*}
Given the regression parameters, the value of the central pixel is estimated as $\hat{y} = \hat{\beta}_0$. Since pixels close to the central pixel are more important, the errors are weighted by a kernel, $K(\mathbf{x},\mathbf{x}_i)$. The estimation problem thus becomes
\begin{align*}
&\min_{\beta_0,\boldsymbol{\beta}_1,\boldsymbol{\beta}_2} \sum_{i=1}^P \left| y_i - \beta_0 - \boldsymbol{\beta}_1^\top (\mathbf{x - x}_i) \right.\\
&\left. - \boldsymbol{\beta}_2^\top \mathrm{vech}( (\mathbf{x - x}_i)(\mathbf{x - x}_i)^\top) \right|^2 K(\mathbf{x},\mathbf{x}_i) , 
\end{align*}
where we used the kernel
\begin{align*}
K(\mathbf{x},\mathbf{x}_i) = \exp \left(- || \mathbf{x - x}_i||_2^2/ r^2 \right) \left( 1 + \mathbf{x^\top x}_i \right)^p , 
\end{align*}
the kernel is a composition of a Gaussian and polynomial kernel \cite{Mitra2,Takeda}. In the simulation we used $r=2.1$ and $p=1$ as in \cite{Mitra2}. To avoid overfitting, it is beneficial to promote sparsity in $[\beta_0, \, \boldsymbol{\beta}_1^\top , \, \boldsymbol{\beta}_2^\top]^\top$ \cite{Elad, Elad2, Elad3}.


We compared the algorithms by computing the Peak Signal to Noise Ratio (PSNR)
\begin{align*}
\text{PSNR} = -10 \cdot \log_{10} \left( \frac{E [ || \mathbf{X - \hat{X}}||_F^2 ] }{E[\max_{i,j} |X_{ij}|^2] (pq)} \right) ,
\end{align*}
where the size of the image is $p \times q$ and the expectation is taken over the different images and realizations of the noise. All images in the simulation were of size $p=q=256$ with $\max_{i,j} |X_{ij}|^2 = 1$. Figure \ref{fig:image_test1} shows one realization of the problem, where SD-RVM gives lower PSNR than the median filter and RB-RVM. In the simulations we varied $\rho$ and used $10$ noise realizations for each image. The result is shown in figure~\ref{fig:image1}.

We found that the median filter performed best for $\rho \leq 0.1$ while SD-RVM outperformed RB-RVM for all values of $\rho$ and also the median filter for $\rho \geq 0.2$. The gain in using SD-RVM over the median filter was significant for the images Boat, Elaine, Lena, House and Peppers (for $\rho \geq 0.2$). The mean cputime of SD-RVM was $68\%$ of the mean cputime of RB-RVM (see Table~\ref{tab:res_image}) while the median filter was by far the fastest method.

\begin{table}
\small
\begin{center}
\caption{Mean cputime (in seconds) for denoising images corrupted by salt and pepper noise averaged over $7$ images.}
\renewcommand{\arraystretch}{1.2}
\begin{tabular}{lcc}
\hline
\textbf{Algorithm} & \textbf{Mean cputime}\\
\hline
Median filter & 5\\
RVM & 925\\
RB-RVM & 1154\\
SD-RVM & 788
\end{tabular}
\end{center}
\vspace*{-0.3cm}
\label{tab:res_image}
\end{table}

%


\section{Conclusion}

In this paper we introduced the combined Sparse and Dense noise Revelance Vector Machine (SD-RVM) which is robust to sparse and dense additive noise. SD-RVM was shown to be equivalent to the minimization of a non-symmetric sparsity promoting cost function.
Through simulations, SD-RVM was shown to empirically perform better than the standard RVM and the robust RB-RVM.

\section{Appendix: Derivation of update equations}

Here we derive the update equations for $\mathbf{\hat{x}}$, $\gamma_i$ and $\beta_j$ for the SD-RVM in Sections \ref{subsec:SD-RVM}, \ref{SD-RVM_blocks} and \ref{unknownblocks}.

For fixed precisions $\boldsymbol{\gamma}$ and $\boldsymbol{\beta}$, the Maximum A Posteriori (MAP) estimate of $\mathbf{x}$ becomes
\begin{align*}
\mathbf{\hat{x}} &= \arg \max_{\mathbf{x}} \, \, \log p(\mathbf{y,x}|\boldsymbol{\gamma},\boldsymbol{\beta})\\
&= \arg \min_{\mathbf{x}} \, \, (\mathbf{y - Ax})^\top \mathbf{B}(\mathbf{y-Ax}) + \mathbf{x}^\top \boldsymbol{\Gamma} \mathbf{x}\\
&= \boldsymbol{\Sigma} \mathbf{A}^\top \mathbf{B} \mathbf{y},
\end{align*}
where $\boldsymbol{\Sigma} = (\boldsymbol{\Gamma} + \mathbf{A}^\top\mathbf{B}\mathbf{A})^{-1}$. The form of the MAP estimate is the same for all models considered in this paper.

\subsection{Derivation of \eqref{eq:xhat_identity} and \eqref{eq:res_identity}}

\label{xhat_identity_derivation}

\begin{proof}[Proof of \eqref{eq:xhat_identity}] Since
\begin{align*}
\mathbf{B}^{-1} + \mathbf{A \boldsymbol{\Gamma}^{-1} A^\top} = \mathbf{B}^{-1} + \sum_{i=1}^n \gamma_i^{-1} \mathbf{a}_i \mathbf{a}_i^\top , 
\end{align*}
where $\mathbf{a}_i$ is the $i$'th column vector of $\mathbf{A}$ we find that
\begin{align*}
&\frac{\partial}{\partial \gamma_i} \left( \mathbf{y}^\top (\mathbf{B}^{-1} + \mathbf{A}\boldsymbol{\Gamma}^{-1} \mathbf{A}^\top)^{-1} \mathbf{y} \right) \\ &=\gamma_i^{-2} \left( \mathbf{a}_i^\top (\mathbf{B}^{-1} + \mathbf{A}\boldsymbol{\Gamma}^{-1} \mathbf{A}^\top)^{-1} \mathbf{y} \right)^2 . 
\end{align*}
Using that
\begin{align}
&\boldsymbol{\Gamma}^{-1}\mathbf{A}^\top(\mathbf{B}^{-1} + \mathbf{A}\boldsymbol{\Gamma}^{-1}\mathbf{A}^\top)^{-1} \mathbf{y}  \label{eq:xhat_derivation}\\ 
&=\boldsymbol{\Gamma}^{-1}\mathbf{A}^\top \left( \mathbf{B} - \mathbf{BA}(\boldsymbol{\Gamma} + \mathbf{A^\top B A})^{-1} \mathbf{A^\top B}\right)\mathbf{y} \nonumber\\
&=\boldsymbol{\Gamma}^{-1}\mathbf{A}^\top \mathbf{B} \mathbf{y} - \boldsymbol{\Gamma}^{-1}\underbrace{\mathbf{A}^\top \mathbf{BA}}_{=\boldsymbol{\Sigma}^{-1} - \boldsymbol{\Gamma}}\underbrace{(\boldsymbol{\Gamma} + \mathbf{A^\top B A})^{-1}}_{= \boldsymbol{\Sigma}} \mathbf{A^\top B}\mathbf{y} \nonumber\\
&= \boldsymbol{\Sigma} \mathbf{A^\top B y} = \mathbf{\hat{x}} , \nonumber
\end{align}
we get that
\begin{align*}
\frac{\partial}{\partial \gamma_i} \left( \mathbf{y}^\top (\mathbf{B}^{-1} + \mathbf{A}\boldsymbol{\Gamma}^{-1} \mathbf{A}^\top)^{-1} \mathbf{y} \right) =  \gamma_i^{-2} \left( \gamma_i \hat{x}_i \right)^2 = \hat{x}_i^2 . 
\end{align*}
\end{proof}

\begin{proof}[Proof of \eqref{eq:res_identity}] Since
\begin{align*}
&\mathbf{y}^\top (\mathbf{B}^{-1} + \mathbf{A \boldsymbol{\Gamma}^{-1} A^\top})^{-1} \mathbf{y}\\
& = \mathbf{y}^\top \mathbf{B} \mathbf{y}
 - \mathbf{y} \mathbf{BA}\underbrace{(\boldsymbol{\Gamma} + \mathbf{A^\top B A})^{-1}}_{=\boldsymbol{\Sigma}} \mathbf{A^\top B y} , 
\end{align*}
we get that
\begin{align*}
&\frac{\partial}{\partial \beta_j} \left( \mathbf{y}^\top (\mathbf{B}^{-1} + \mathbf{A}\boldsymbol{\Gamma}^{-1} \mathbf{A}^\top)^{-1} \mathbf{y} \right) \\
&=y_j^2 - 2 y_j \mathbf{A}_{j,:} \boldsymbol{\Sigma} \mathbf{A^\top B y} + \mathbf{y}^\top \mathbf{BA} \boldsymbol{\Sigma} \mathbf{A}_{j,:}^\top \mathbf{A}_{j,:}  \boldsymbol{\Sigma} \mathbf{A^\top B y}\\
&= y_j^2 - 2y_j \mathbf{A}_{j,:} \mathbf{\hat{x}} + (\mathbf{A}_{j,:} \mathbf{x})^2 = [\mathbf{y - A\hat{x}}]_j^2 . 
\end{align*}

\end{proof}

\subsection{Known block structure}

\label{sec:derivation_SDRVM_knownblocks}

Let $\boldsymbol{\Gamma}$ and $\mathbf{B}$ be diagonal matrices with
\begin{align*}
&[\boldsymbol{\Gamma}]_{kk} = \gamma_i  \text{, if } k \in I_i, 
&[\mathbf{B}]_{ll} = \beta_j  \text{, if } l \in J_j , 
\end{align*}
and zero otherwise.

To update the precisions we maximize the marginal distribution
\begin{align*}
p(\mathbf{y},\boldsymbol{\gamma}, \mathbf{B}) = p(\mathbf{y}|\boldsymbol{\gamma}, \boldsymbol{\beta})p(\boldsymbol{\gamma})p(\boldsymbol{\beta}) , 
\end{align*}
with respect to $\boldsymbol{\gamma}$ and $\boldsymbol{\beta}$, where $p(\boldsymbol{\gamma})$ and $p(\boldsymbol{\beta})$ is as in \eqref{alphaprior} and \eqref{betaprior}. The log-likelihood of the parameters is
\begin{align*}
&\mathcal{L} = \text{const.} - \frac{1}{2} \log \mathrm{det}(\mathbf{B}^{-1} + \mathbf{A} \boldsymbol{\Gamma}^{-1} \mathbf{A^\top}) \nonumber\\
&- \frac{1}{2} \mathbf{y}^\top (\mathbf{B}^{-1} + \mathbf{A}\boldsymbol{\Gamma}^{-1} \mathbf{A}^\top)^{-1} \mathbf{y} \nonumber\\
& + \sum_{i=1}^p (a\log \gamma_i - b\gamma_i) + \sum_{j=1}^q (c\log \beta_j - d\beta_j) . 
\end{align*}
Using \eqref{determinantlemma} we get that $\mathcal{L}$ is maximized when
\begin{align}
&\frac{\partial \mathcal{L}}{\partial \gamma_i} = - \frac{1}{2} \mathrm{tr}(\boldsymbol{\Sigma}_{I_i}) + \frac{n_i}{2\gamma_i} + \frac{a}{\gamma_i} - b \nonumber\\
&-\frac{1}{2\gamma_i^2}||\mathbf{A}_{I_i}^\top(\mathbf{B}^{-1} + \mathbf{A}\boldsymbol{\Gamma}^{-1}\mathbf{A}^\top)^{-1} \mathbf{y} ||_2^2  = 0 , 
\label{derL}
\end{align}
where $\boldsymbol{\Sigma}_{I_i} \in \mathbb{R}^{n_i \times n_i}$ is the submatrix of $\boldsymbol{\Sigma}$ consisting of the columns and rows in $I_i$. Further, using \eqref{eq:xhat_derivation} we get that
\begin{align}
\label{xhatid}
\mathbf{A}_{I_i}^\top(\mathbf{B}^{-1} + \mathbf{A}\boldsymbol{\Gamma}^{-1}\mathbf{A}^\top)^{-1} \mathbf{y} = \gamma_i \mathbf{\hat{x}}_{I_i} . 
\end{align}
Thus, \eqref{derL} is fulfilled when
\begin{align*}
 - \frac{1}{2} \mathrm{tr}(\boldsymbol{\Sigma}_{I_i}) + \frac{n_i}{2\gamma_i} + \frac{a}{\gamma_i} - b - \frac{1}{2} ||\mathbf{\hat{x}}_{I_i}||_2^2 = 0 . 
\end{align*}
As before, instead of solving for $\gamma_i$ we rewrite the equation as
\begin{align}
\label{alphanew}
n_i - \gamma_i \mathrm{tr}(\boldsymbol{\Sigma}_{I_i}) + 2a - (||\mathbf{\hat{x}}_{I_i}||_2^2 + 2b) \gamma_i^{new} = 0 . 
\end{align}
Solving \eqref{alphanew} for $\gamma_i^{new}$ gives us the update equation \eqref{blockgamma_update}.

To find the update equation for $\beta_j$ we use that
\begin{align*}
&\frac{\partial}{\partial \beta_j} \left[ \mathbf{y}^\top (\mathbf{B}^{-1} +\mathbf{A}\boldsymbol{\Gamma}^{-1} \mathbf{A}^\top)^{-1} \mathbf{y} \right]
= ||\mathbf{y}_{J_j}||_2^2 \\
&- 2\mathbf{y}_{J_j}^\top \mathbf{A}_{J_j,:} \boldsymbol{\Sigma} \mathbf{A}^\top \mathbf{B} \mathbf{y} 
+ \mathbf{y}^\top \mathbf{B}\mathbf{A} \boldsymbol{\Sigma} \mathbf{A}_{J_j,:}^\top \mathbf{A}_{J_j,:}  \boldsymbol{\Sigma} \mathbf{A}^\top \mathbf{B} \mathbf{y} \\&
= ||(\mathbf{y - A\hat{x}})_{J_j}||_2^2, 
\end{align*}
where $\mathbf{A}_{J_j,:}$ consists of the row vectors of $\mathbf{A}$ which row number belongs to $J_j$. We get that
\begin{align*}
&\frac{\partial \mathcal{L}}{\partial \beta_j} = - \frac{1}{2} \mathrm{tr}(\boldsymbol{\Sigma} \mathbf{A}_{J_j,:}^\top \mathbf{A}_{J_j,:}) + \frac{m_j}{2\beta_j} \\& - \frac{1}{2} ||(\mathbf{y - A\hat{x}})_{J_j}||_2^2 + \frac{c}{\beta_j} - d = 0 . 
\end{align*}
Rewriting the equation as
\begin{align*}
&1 - \beta_j \mathrm{tr}(\mathbf{A}_{J_j,:} \boldsymbol{\Sigma} \mathbf{A}_{J_j,:}^\top) + 2c \\
&- (||(\mathbf{y - A\hat{x}})_{J_j}||_2^2 + 2d) \beta_j^{new} = 0 , 
\end{align*}
and using that $\mathrm{tr}(\mathbf{A}_{J_j,:} \boldsymbol{\Sigma} \mathbf{A}_{J_j,:}^\top) = \mathrm{tr}([\mathbf{A}\boldsymbol{\Sigma}\mathbf{A}^\top]_{J_j})$ gives us the update equation \eqref{blockbetaupdate}.

\subsection{Unknown block structure}

When the block structure is unknown, we use the overparametrized model in section \ref{unknownblocks}. The log-likelihood of the parameters is
\begin{align}
&\mathcal{L} = \log p(\mathbf{y}|\boldsymbol{\gamma}, \boldsymbol{\beta})p(\boldsymbol{\gamma})p(\boldsymbol{\beta}) \nonumber \\
&=\text{const.} - \frac{1}{2} \log \mathrm{det}(\mathbf{B}^{-1} + \mathbf{A} \boldsymbol{\Gamma}^{-1} \mathbf{A^\top}) \label{loglikelihood}\\
&- \frac{1}{2} \mathbf{y}^\top (\mathbf{B}^{-1} + \mathbf{A}\boldsymbol{\Gamma}^{-1} \mathbf{A}^\top)^{-1} \mathbf{y} \nonumber\\
& + \sum_{i=1}^p (a\log \tilde{\gamma}_i - b\tilde{\gamma}_i) + \sum_{j=1}^q (c\log \tilde{\beta}_j - d\tilde{\beta}_j)  . \nonumber
\end{align}

We search to maximize \eqref{loglikelihood} with respect to the underlying variables $\tilde{\gamma}_k$ and $\tilde{\beta}_l$. Using that $\frac{\partial \gamma_i^{-1}}{\partial \tilde{\gamma}_k}  = - \tilde{\gamma}_k^{-2}$, $\frac{\partial \gamma_i}{\partial \tilde{\gamma}_k} = \gamma_i^2 \tilde{\gamma}_k^{-2}$, when $i \in I_k$ and zero otherwise, \eqref{determinantlemma} and \eqref{eq:xhat_derivation} we find that $\mathcal{L}$ is maximized when
\begin{align}
&\frac{\partial \mathcal{L}}{\partial \tilde{\gamma}_k} = - \frac{1}{2\tilde{\gamma}_k^2} \mathrm{tr}(\boldsymbol{\Sigma} \boldsymbol{\Gamma}_k^2) + \frac{1}{2\tilde{\gamma}_k^2} \mathrm{tr}(\boldsymbol{\Gamma}_k) \nonumber \\
&- \frac{1}{2 \tilde{\gamma}_k^2} \mathbf{\hat{x}}^\top \boldsymbol{\Gamma}_k^2 \mathbf{\hat{x}} + \frac{a}{\tilde{\gamma}_k} - b = 0 , \label{derivativezero}
\end{align}
By rewriting \eqref{derivativezero} as
\begin{align}
&\frac{1}{\tilde{\gamma}_k} \mathrm{tr}(\boldsymbol{\Gamma}_k) - \frac{1}{\tilde{\gamma}_k} \mathrm{tr}(\boldsymbol{\Gamma}_k \boldsymbol{\Sigma} \boldsymbol{\Gamma}_k) + 2a \nonumber\\
&- \left( \frac{1}{\tilde{\gamma}_k^2} ||\boldsymbol{\Gamma}_k \mathbf{\hat{x}}||_2^2 + 2b\right) \tilde{\gamma}_k^{new} = 0 . \label{derivativezero2}
\end{align}
Solving \eqref{derivativezero2} for $\tilde{\gamma}_k^{new}$ gives us the update equation \eqref{gamma_unknown_update}.

For the noise precisions, we similarly find that
\begin{align*}
&\frac{\partial \mathcal{L}}{\partial \tilde{\beta}_l} = - \frac{1}{2\tilde{\beta}_l^2}\mathrm{tr}(\mathbf{B}_l \mathbf{A}^\top \boldsymbol{\Sigma} \mathbf{A}\mathbf{B}_l) + \frac{1}{2\tilde{\beta}_l^2} \mathrm{tr}(\mathbf{B}_l) \\
& - \frac{1}{2\tilde{\beta}_l^2} ||\mathbf{B}_l(\mathbf{y - A\hat{x}})||_2^2 + \frac{c}{\tilde{\beta}_l} - d = 0 .
\end{align*}
By rewriting the expression as
\begin{align*}
&\frac{1}{\tilde{\beta}_l} \mathrm{tr}(\mathbf{B}_l) - \frac{1}{\tilde{\beta}_l} \mathrm{tr}(\mathbf{B}_l \mathbf{A}^\top \boldsymbol{\Sigma} \mathbf{A}\mathbf{B}_l) + 2c\\
&- \left( \frac{1}{\tilde{\beta}_l^2} ||\mathbf{B}_l(\mathbf{y - A\hat{x}})||_2^2 + 2d \right)\tilde{\beta}_l^{new} = 0 , 
\end{align*}
we find the update equation \eqref{beta_unknown_update}.

We see that the form of update equations depends on how the equations are rewritten. The form used here has the advantage of reducing to \eqref{blockbetaupdate} when the underlying blocks are disjoint.

%









\begin{thebibliography}{1}


\bibitem{Tipping1}
M.~Tipping, \emph{The relevance vector machine}, NIPS, 1999, pp. 652-658.

\bibitem{Bishop}
C.~Bishop, \emph{Pattern Recognition and Machine Learning}, Springer-Verlag New York, Inc. Secaucus, NJ, USA, 2006.

\bibitem{Wipf}
D.P.~Wipf and B.D.~Rao, \emph{Sparse Bayesian learning for basis selection}, IEEE Transactions on Signal Processing, vol.52, no.8, pp.2153 - 2164, Aug. 2004.

\bibitem{Wipf2}
D.~Wipf, J.~Palmer and B.D.~Rao, \emph{Perspectives on sparse Bayesian learning}, Advances in neural information processing systems, vol. 16, pp. 249 - 256, 2004.

\bibitem{BayesianCS}
S.~Ji, Y.~Xue and L.~Carin, \emph{Bayesian compressive sensing}, IEEE Transactions on Signal Processing, vol. 56, no. 6, pp. 2346-2356, 2008.

\bibitem{Mitra2}
K.~Mitra, A.~Veeraraghavan and R.~Chellappa, \emph{Robust RVM regression using sparse outlier model}, 2012 IEEE Conference on Computer Vision and Pattern Recognition (CVPR), 2012, pp. 1887-1894.

\bibitem{Laska}
J.~Laska, M.~Davenport and R.~Baraniuk, \emph{Exact signal recovery from sparsely corrupted measurements through the pursuit of justice}, Proceedings of the 43rd Asimolar conference on Signals, systems and computers, Piscataway, NJ, USA, 2009, pp. 1556-1560, IEEE Press.

\bibitem{Jin}
Y.~Jin and B.D.~Rao, \emph{Algorithms for robust linear regression by exploiting the connection to sparse signal recovery}, IEEE International Conference on Acoustics Speech and Signal Processing (ICASSP), 2010, pp.3830 - 3833, 14-19 March 2010.

\bibitem{Vehkapera}
M.~Vehkapera, Y.~Kabashima and S.~Chatterjee, \emph{Statistical mechanics approach to sparse noise denoising}, Proceedings of the 21st European Signal Processing Conference (EUSIPCO), 2013, pp. 1-5, 9-13 September 2013.

\bibitem{Cherian}
A.~Cherian, S.~Sra and N.~Papanikolopoulos, \emph{Denoising sparse noise via online dictionary learning}, IEEE International Conference on Acoustics, Speech and Signal Processing (ICASSP), 2011, pp. 2060 - 2063, 22-27 May 2011.

\bibitem{Giri}
R.~Giri and B.D.~Rao, \emph{Block sparse excitation based all-pole modeling of speech}, IEEE International Conference on Acoustics, Speech and Signal Processing (ICASSP), 2014, pp. 3754 - 3758, 4-9 May 2014.

\bibitem{Giacobello}
D.~Giacobello, M.G.~Christensen, M.N.~Murthi, S.H.~Jensen and M.~Moonen, \emph{Sparse Linear Prediction and Its Applications to Speech Processing}, IEEE Transactions on Audio, Speech and Language Processing, vol.20, no.5, pp.1644 - 1657, July 2012.

\bibitem{Kekatos}
V.~Kekatos and G.B.~Giannakis, \emph{From Sparse Signals to Sparse Residuals for Robust Sensing}, IEEE Transactions on Signal Processing, vol.59, no.7, pp. 3355 - 3368, July 2011.

\bibitem{Carrillo}
R.E.~Carrillo, K.E.~Barner and T.C.~Aysal, \emph{Robust Sampling and Reconstruction Methods for Sparse Signals in the Presence of Impulsive Noise}, IEEE Journal of Selected Topics in Signal Processing, vol.4, no.2, pp. 392 - 408, April 2010.

\bibitem{Wright}
J.~Wright and Y.~Ma, \emph{Robust face recognition via sparse representation}, IEEE Transactions on Information Theory, vol. 56, no. 7, pp. 3540-3560, July 2010.

\bibitem{zhang}
Z.~Zhang and B.D.~Rao, \emph{Extension of sbl algorithms for the recovery of block sparse signals with intra-block correlation}, IEEE Transactions on Signal Processing, vol. 61, no. 8, pp. 2009-2015, April 2013.

\bibitem{zhang2}
Z.~Zhang and B.D.~Rao, \emph{Sparse Signal Recovery With Temporally Correlated Source Vectors Using Sparse Bayesian Learning}, IEEE Journal of Selected Topics in Signal Processing, vol.5, no.5, pp.912,926, Sept. 2011.

\bibitem{Chen2001}
S.~Chen, D.~Donoho and M.~Saunders, \emph{Atomic decomposition by basis pursuit}, SIAM Rev., vol. 43, no. 1, pp. 129-159, January 2001.

\bibitem{Lookahead}
S.~Chatterjee, D.~Sundman, M.~Vehkapera, M.~Skoglund, \emph{Projection-Based and Look-Ahead Strategies for Atom Selection}, IEEE Transactions on Signal Processing, vol.60, no.2, pp.634 - 647, Feb. 2012.

\bibitem{DIP}
D.~Zachariah, S.~Chatterjee, M.~Jansson, \emph{Dynamic Iterative Pursuit}, IEEE Transactions on Signal Processing, vol.60, no.9, pp.4967 - 4972, Sept. 2012.

\bibitem{harville08}
D.~Harville, \emph{Matrix algebra from a statistician's perspective}, Springer, 2008.

\bibitem{Mackay}
D.~MacKay, \emph{Bayesian interpolation}, Neural Computation, vol. 4, pp. 415-447, 1991.

\bibitem{Rojas}
C.~Rojas, D.~Katselis and H.~Hjalmarsson, \emph{A note on the spice method}, IEEE transactions on signal processing, vol 61. no. 18, pp. 4545-4551, 2013.

\bibitem{num_linalg}
L.~Trefethen and D.~Bau, \emph{Numerical linear algebra}, Society for Industrial and Applied Mathematics, 1997.

\bibitem{boston}
K.~Bache and M.~Lichman, \emph{UCI machine learning repository}, 2013.

\bibitem{cvx}
M.~Grant and S.~Boyd, \emph{CVX: Matlab Software for Disciplined Convex Programming, version 2.1}, http://cvxr.com/cvx, March 2014.

\bibitem{Candes2}
E.~Candes, J.~Romberg and T.~Tao, \emph{Stable signal recovery from incomplete and inaccurate measurements}, Communications on Pure and Applied Mathematics, vol. 59, no. 8, pp. 1207-1223, 2006.

\bibitem{modelCS}
R.G.~Baraniuk, V.~Cevher, M.F.~Duarte, C.~Hegde, \emph{Model-Based Compressive Sensing}, IEEE Transactions on Information Theory, vol.56, no.4, pp. 1982 - 2001, April 2010.

\bibitem{Takeda}
H.~Takeda, S.~Farsiu and P.~Milanfar, \emph{Robust kernel regression for restoration and reconstruction of images from sparse noisy data}, IEEE International Conference on Image Processing, 2006.



\bibitem{Elad}
M.~Elad and M.~Aharon, \emph{Image Denoising Via Sparse and Redundant Representations Over Learned Dictionaries}, IEEE Transactions on Image Processing, vol.15, no.12, pp. 3736 - 3745, Dec. 2006.

\bibitem{Elad2}
J.~Mairal, M.~Elad and G.~Sapiro, \emph{Sparse Representation for Color Image Restoration}, IEEE Transactions on Image Processing, vol.17, no.1, pp. 53 - 69, Jan. 2008.

\bibitem{Elad3}
A.~Bruckstein, D.~Donoho and M.~Elad, \emph{From Sparse Solutions of Systems of Equations to Sparse Modeling of Signals and Images}, SIAM Review, vol.51, no.1, pp. 34 - 81, 2009.




\end{thebibliography}
\end{document}